\documentclass[10pt,twocolumn,letterpaper]{article}

\usepackage[pagenumbers]{cvpr} 
\usepackage{booktabs}
\usepackage{tabularx}
\usepackage[most]{tcolorbox}
\usepackage{microtype}

\usepackage{cuted}

\definecolor{cvprblue}{rgb}{0.21,0.49,0.74}
\usepackage[pagebackref,breaklinks,colorlinks,allcolors=cvprblue]{hyperref}

\def\confName{CVPR}
\def\confYear{2027}

\title{ProxyGenBench: Benchmarking Proxy-Guided Video Generation}

\author{
\begin{tabular}{c}
Zhenchen Tang\textsuperscript{1,4,5,*}, Haolin Lu\textsuperscript{2,4,*},
Songlin Yang\textsuperscript{3,4}, Xiaotong Zhao\textsuperscript{4}, Ran Zhang\textsuperscript{4},\\
Bo Peng\textsuperscript{1,5}, Shuai Li\textsuperscript{4},
Haotian Fan\textsuperscript{4}, Haobin Zhong\textsuperscript{4}, Yanni Zhang\textsuperscript{4},\\
Songlian Li\textsuperscript{4}, Ning Zhang\textsuperscript{4},
Xun Gao\textsuperscript{4}, Alan Zhao\textsuperscript{4}, Jing Dong\textsuperscript{1,5,$\dagger$}\\[4pt]
\small \textsuperscript{1}New Laboratory of Pattern Recognition, Institute of Automation,\\
\small Chinese Academy of Sciences\\
\small \textsuperscript{2}University of California, San Diego\quad
\textsuperscript{3}The Hong Kong University of Science and Technology\\
\small \textsuperscript{4}Tencent\quad
\textsuperscript{5}School of Artificial Intelligence, University of Chinese Academy of Sciences\\[3pt]
\small \textsuperscript{*}Equal contribution.\quad
\textsuperscript{$\dagger$}Corresponding author.\\[3pt]
\small \textbf{Project page:} \url{https://2kxx.github.io/ProxyGenBench/}
\end{tabular}
}

\begin{document}
\maketitle
\begin{strip}
\centering
\includegraphics[width=\linewidth]{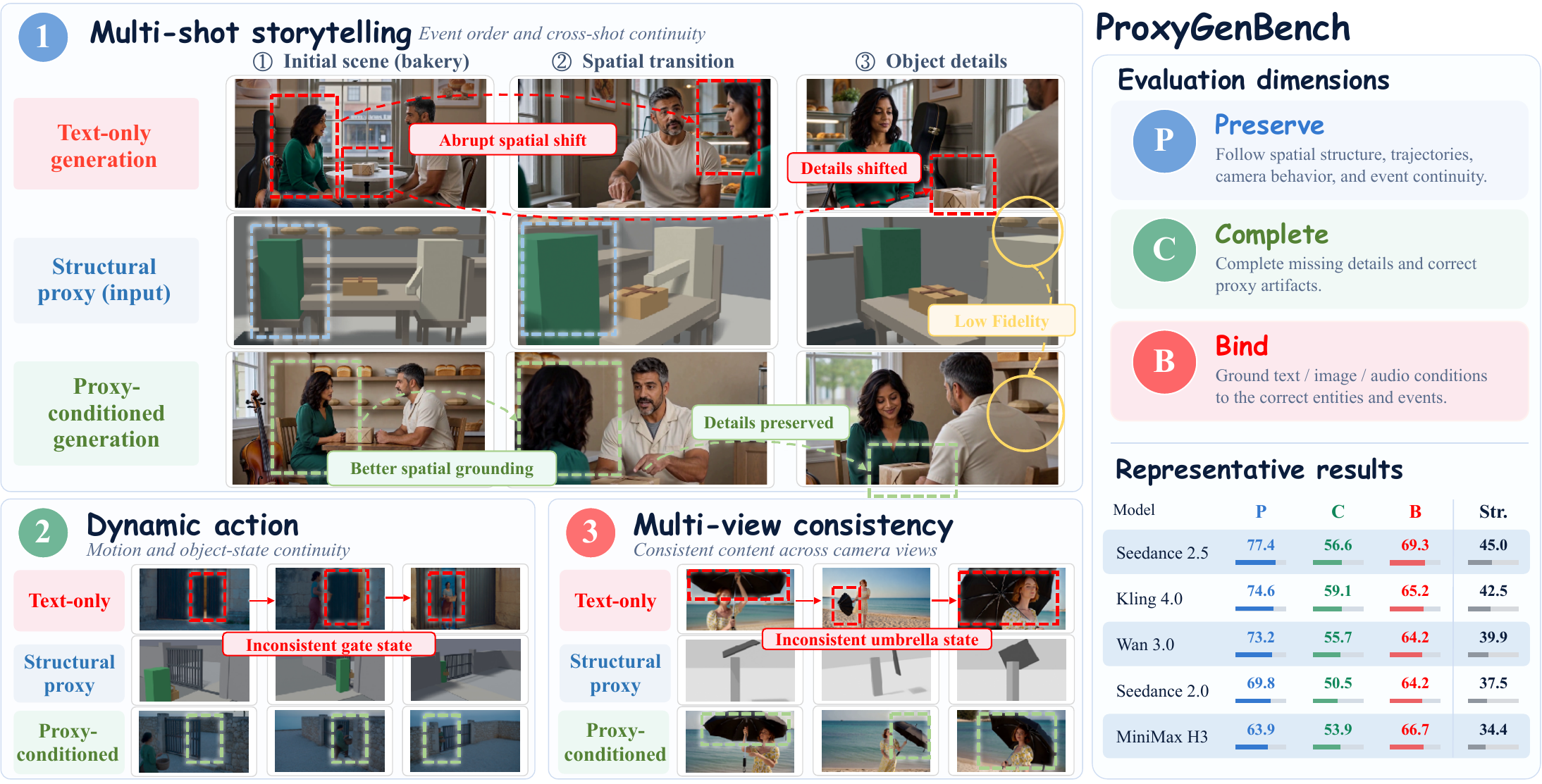}
\captionsetup{hypcap=false}
\captionof{figure}{\textbf{Overview of ProxyGenBench.} The \textsc{Bakery} example shows how proxy guidance improves spatial and object-detail consistency across shots, while the \textsc{DynamicAction} and \textsc{MultiView} examples highlight consistency in gate and umbrella states. The proxies supply coarse scene structure rather than complete target videos. The right panel summarizes \textit{Preserve}, \textit{Complete}, and \textit{Bind} dimensions and reports rounded scores from \cref{tab:overall}(a) on a 0--100 scale; Str. denotes our \textit{Strict} score, a structure-aware aggregation of all three dimensions.}
\captionsetup{hypcap=true}
\label{fig:teaser}%
\end{strip}
\begin{abstract}
Recent video generators produce realistic videos, yet controlling how scenes unfold remains challenging. Structural proxies, such as animated 3D blockouts, guide generation by specifying scene layout, motion, camera behavior, and event order. However, these coarse inputs leave details unspecified and may contain artifacts that should not be reproduced. Successful proxy-guided generation must therefore \textbf{Preserve} intended structure, \textbf{Complete} missing content and correct proxy artifacts, and \textbf{Bind} text, image, and audio conditions to the right entities and events. We introduce \textbf{ProxyGenBench} to study these interacting requirements, with 312 evaluation cases from 60 scenes across three scene tasks and controlled variations in proxy fidelity, text detail, and multimodal inputs. We evaluate five video generators using verified, case-specific checklists. Experiments show that proxies improve spatiotemporal control, but greater proxy detail can coincide with poorer completion, while strong structural adherence does not ensure correct binding.
\end{abstract}
\section{Introduction}
\label{sec:introduction}

Video generation has advanced rapidly in realism, but precise control over how a scene unfolds remains challenging~\cite{yang2025cogvideox, polyak2024movie, bar2024lumiere}. Generated videos may contain incorrect interactions, inconsistent spatial relationships, or discontinuities across shots, even when individual frames appear plausible~\cite{ren2024consisti2v,an2026onestory}. Text prompts and reference images provide semantic and appearance guidance, but often leave object trajectories, camera behavior, and event timing implicit~\cite{he2025cameractrl, wang2024motionctrl}. Structural proxies, such as animated 3D blockouts, complement these inputs by representing scene layout, motion, camera behavior, and event order in a coarse video sequence.


While structural proxies offer more explicit control over scene organization, their guidance is not without limitations, as illustrated in Fig.~\ref{fig:teaser}. In the \textit{bakery} scene, text-only generation introduces an abrupt spatial shift and changes object details across shots, while proxy guidance helps preserve scene layout and object continuity. 
Similar improvements are observed in the \textit{dynamic-action} and \textit{multi-view} examples.
However, proxy guidance often contains low-fidelity geometry and motion that may mislead the generator and hinder realistic scene realization, as illustrated by the low-poly donuts in the \textit{bakery} example.
What should be preserved from the proxy and what should be refined ultimately depends on the user's intent, making additional conditions, such as text and reference images, essential for guiding the generation.
This raises a central question: \textit{how can generators preserve useful proxy constraints without inheriting their artifacts or overlooking information from other conditions?} Existing benchmarks provide limited insight into how these requirements interact in proxy-guided generation.

To address this gap, we identify three complementary abilities essential for successful proxy-guided generation: \textbf{Preserve} intended structure, \textbf{Complete} missing content and correct proxy artifacts, and \textbf{Bind} additional conditions to the appropriate entities and events. To evaluate these abilities, we introduce \textbf{ProxyGenBench}, comprising 312 evaluation cases from 60 scenes across dynamic action, multi-shot storytelling, and multi-view consistency challenges.


Our scenes cover four potential applications of proxy-guided generation: enhancing reference-to-video workflows, enabling code-driven proxy authoring, supporting game-like world modeling, and assisting 3D animation completion. For each scene, we provide multiple variants to study how models respond to different levels of conditioning detail and balance guidance across modalities. Low-fidelity proxies leave more room for refinement but may provide insufficient structural guidance, while high-fidelity proxies are easier to follow but risk transferring unwanted details. Similarly, simple text prompts may underspecify user intent, whereas detailed prompts or additional image references may distract models from the proxy. Through matched and targeted configurations, we examine when proxies improve control, how proxy fidelity affects completion, and whether additional conditions are correctly integrated into the generated scene.

To evaluate model performance across these variations, we organize 32 fine-grained criteria into eight capabilities under Preserve, Complete, and Bind, and instantiate applicable criteria as verified, case-specific checklist questions. Each checklist is grounded in the inputs available to the generator and fixed across models for the same case. VLM judgments use visual or audio evidence to assess structural adherence, completion, and binding beyond proxy resemblance. We report the three dimension scores alongside a structure-sensitive \textit{Strict} score, which gives additional emphasis to critical Preserve constraints because structural control is the basic requirement of proxy-guided generation.

Experiments on five video generators show that proxy guidance improves spatiotemporal control, particularly action transitions and cross-shot spatial consistency. However, a high average Preserve score does not guarantee that all critical structural constraints are satisfied together. In scene-matched comparisons, moving from coarse to detailed proxies increases Preserve but lowers Complete for all five models, revealing a preservation--completion tension. Results on Studio 3D proxies also show that strong structural adherence can coexist with incorrect condition binding. Together, these findings highlight the need to preserve intended structure selectively, correct proxy artifacts, and ground supplementary information within the same generated scene.

Our contributions are threefold:
\begin{itemize}
    \item \textbf{Benchmark.} We build controlled proxy-guided generation cases varying proxy detail and text, image, and audio requirements.
    \item \textbf{Evaluation.} We develop verified, case-specific checklists to assess intended structural constraints, completion of unfinished content, correction of proxy artifacts, and grounding of additional conditions.
    \item \textbf{Analysis.} We quantify proxy benefits and characterize joint structural, completion, and condition-binding challenges across input variations.
\end{itemize}
\section{Related Work}

\subsection{Controllable Video Generation}
Video generation has increasingly adopted spatial and temporal conditions to complement text and reference images. VideoComposer~\cite{wang2023videocomposer} combines multiple control signals, Motion Prompting~\cite{geng2024motionprompting} uses trajectories to guide scene motion, and SymphoMotion~\cite{zhang2026symphomotion} jointly controls camera and object motion. VideoDirectorGPT~\cite{lin2023videodirectorgpt} uses LLM-guided planning to maintain layout and entity consistency across multiple scenes. 3D Scene Prompting~\cite{lee20253dscenepromptingsceneconsistent} uses scene geometry to maintain consistency across viewpoints. More directly related, Proxy2World~\cite{xu2026proxy2worldlearninggenerateworlds} explores synthesizing detailed videos from lightweight structural proxies and the balance between structural adherence and visual quality. These methods demonstrate the value of spatiotemporal guidance. Our ProxyGenBench instead studies structural inputs that leave content unspecified or contain artifacts requiring correction.

\subsection{Video Generation Benchmarks}
Existing benchmarks assess complementary aspects of generated videos. VBench~\cite{huang2023vbench} and EvalCrafter~\cite{liu2024evalcrafter} evaluate visual quality, temporal consistency, and semantic alignment; VBench-2.0~\cite{zheng2025vbench20advancingvideogeneration} and VideoPhy~\cite{bansal2025videophy} extend evaluation toward physical realism and commonsense. VideoScore~\cite{he2024videoscore} and EvalVerse~\cite{yang2026evalverse} further assess human-aligned and cinematic quality. T2V-CompBench~\cite{sun2025t2v} examines compositional requirements such as attribute binding and object interactions, while TC-Bench~\cite{feng2024tc} focuses on temporal state and relation changes. Concurrent work PROWBench~\cite{huang2026prowbenchvideomodelsrender} evaluates fidelity to program-specified world states and events, including proxy-conditioned generation. ProxyGenBench instead focuses on imperfect structural proxies, requiring models to preserve intended spatial and temporal relationships, correct proxy artifacts, and bind supplementary conditions within the same scene. Varying proxy detail and complementary inputs lets us examine how these requirements interact.

\section{ProxyGenBench}

ProxyGenBench evaluates how different input conditions affect video generation. It tests whether models can preserve the intended scene structure, complete missing content and correct proxy artifacts, and integrate additional text, image, and audio conditions. These challenges often arise together, but a model may perform well on some while struggling with others.

\subsection{Problem Setting and Structural Proxies}

\paragraph{Structural proxy.} 
A structural proxy is a video that specifies the intended scene layout, object placement, motion, camera behavior, and event order without fully defining the final result. Our proxies are rendered from 3D representations obtained from various sources. They may contain placeholder geometry, incomplete interactions, or unnatural motion. We use \emph{proxy fidelity} to describe the level of geometric and motion detail, rather than content accuracy. A higher-fidelity proxy is thus not necessarily more accurate.

\paragraph{Generation setting.} We define each evaluation case as $c=(s,p,U)$, where $s$ is the intended scene specification, $p$ is a structural proxy, and $U$ includes a text prompt and optional image or audio references. Given $(p,U)$, the model generates a video $v$. The scene specification $s$ is used only for benchmark evaluation and is not provided to the video model. We derive evaluation checklists only from the conditions available to the model, ensuring that it is not evaluated on information it never received. Details are in the appendix.

\subsection{Benchmark Construction}

We construct the benchmark by varying several aspects of proxy-guided video generation. Scene tasks define the spatial and temporal relationships to preserve. Proxy sources and fidelity levels determine how much scene structure is provided and what remains incomplete. Additional text, image, and audio inputs introduce further requirements for the same scene. These variations test different aspects of the same generation problem rather than separate tasks.

\paragraph{Scene tasks.} Dynamic action tests motion, interactions, and state transitions; multi-shot storytelling tests event order and continuity across cuts; and multi-view consistency tests stable scene and object states across viewpoints. These tasks cover motion, cross-shot continuity, and cross-view consistency; 
each task is also evaluated on the completion and reference-grounding requirements that apply to it. We additionally annotate five content categories---Characters, Navigation, Objects, Combat \& Transform, and Camera---to describe coverage rather than create separate tasks.

\begin{figure}
    \centering
    \includegraphics[width=\linewidth]{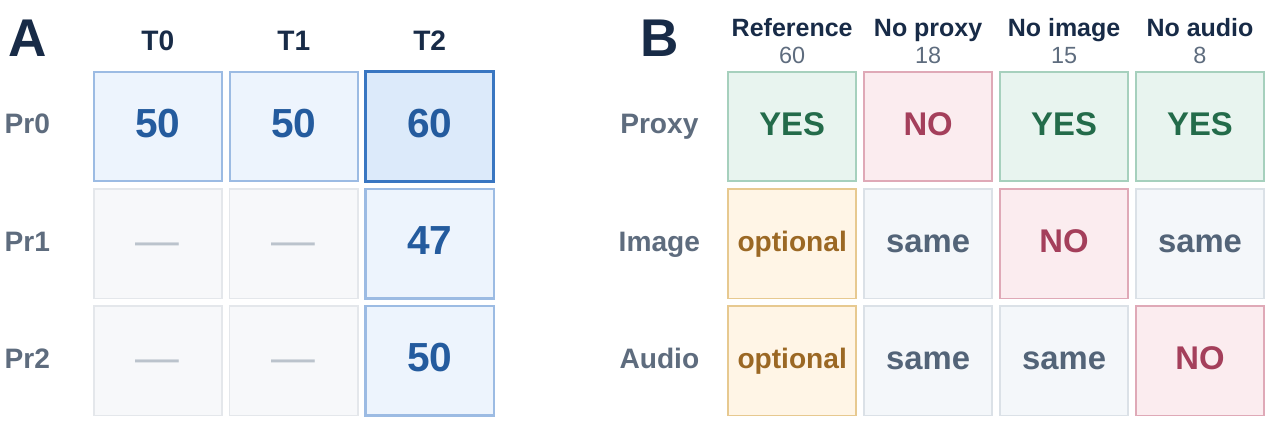}  
    \caption{Evaluation case construction. (a) The proxy--text grid contains 257 cases; dashes mark omitted configurations. (b) Ablations add 18 no-proxy, 15 no-image, and 8 no-audio cases; the 60 reference cases are included in (a). The 14 targeted cases complete the benchmark: 312 cases overall, including 294 with proxy videos.}
    \label{fig:cases}
    \vspace{-0.5cm}
\end{figure}

\begin{figure*}[t]
    \centering
    \includegraphics[width=\linewidth]{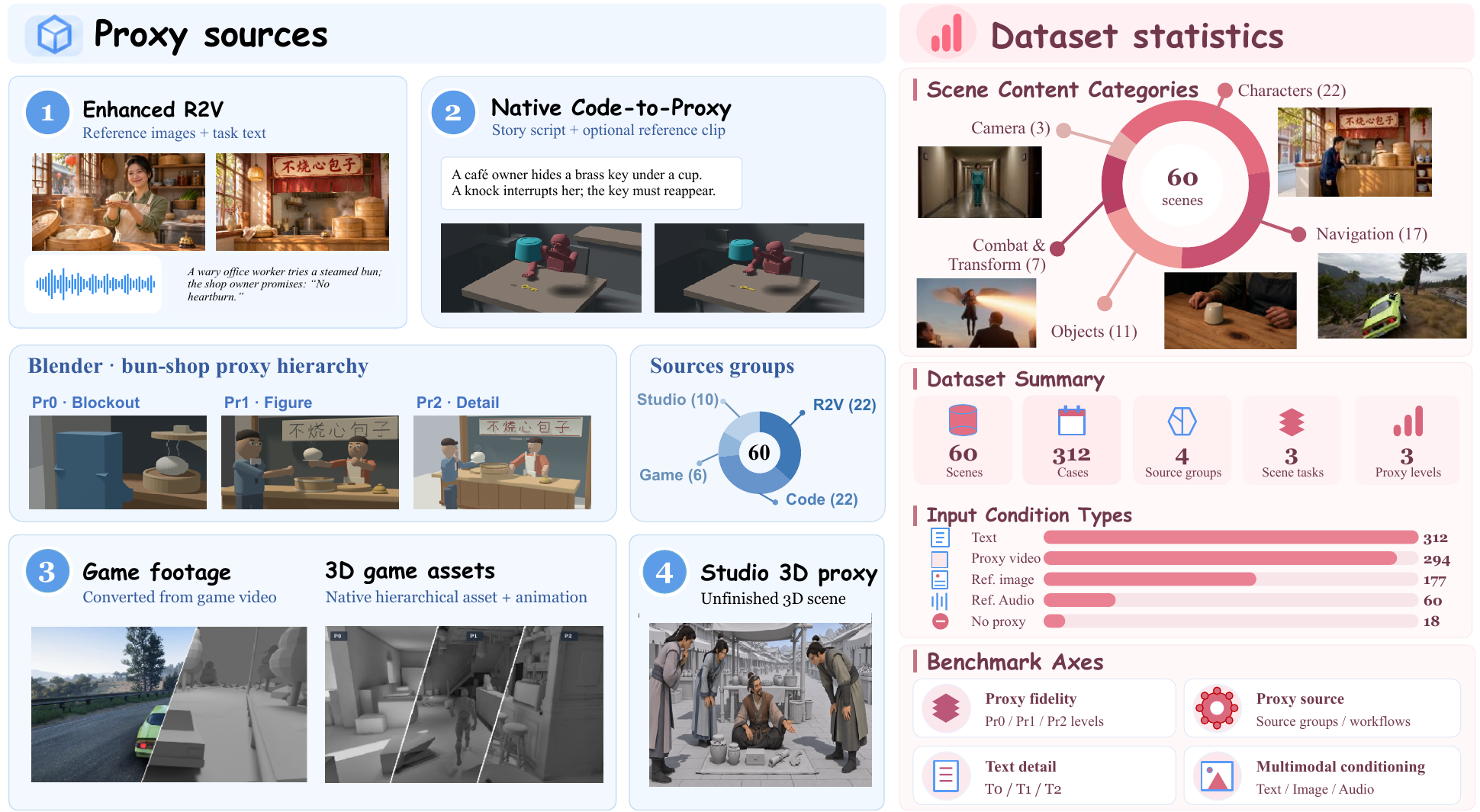}
    \caption{\textbf{Proxy sources and benchmark coverage}: 60 scenes and 312 cases from four source groups. Game Proxy includes game footage and 3D game assets. Source-group and content-category counts are scene-level; input-condition counts are case-level and non-exclusive.}
    \vspace{-0.5cm}
    \label{fig:dataset}
\end{figure*}

\paragraph{Proxy sources.} We collect proxies from four source groups, each with a different construction workflow (Fig.~\ref{fig:dataset}). Enhanced R2V adds proxy guidance to existing image- or audio-conditioned tasks. Native Code-to-Proxy uses an LLM-based agent to animate 3D scenes from scripts or reference videos. Game Proxy includes game footage and 3D game assets, while Studio 3D Proxy uses unfinished production assets. Together, these sources cover a range of incomplete geometry and motion, with different types of artifacts across sources. We provide details of each construction workflow and its associated proxy artifacts in the appendix.

\paragraph{Input variations.} We create multiple evaluation cases from each scene by varying the input conditions. We define three proxy-fidelity levels---Pr0 (Blockout), Pr1 (Figure), and Pr2 (Detail)---with increasing geometric and motion detail. Comparing these levels allows us to study whether more proxy detail improves both structural adherence and natural completion. Text-detail levels T0--T2 vary the amount of semantic guidance. Optional image and audio references add requirements for appearance, voice, and events. We vary text detail at a fixed proxy level (Pr0) and proxy fidelity at a fixed text level (T2) (Fig.~\ref{fig:cases}). This allows us to study their effects separately without testing every combination. 

\paragraph{Evaluation cases.} We construct 312 cases from 60 scenes (Fig.~\ref{fig:cases}), including 294 proxy-conditioned cases and 18 no-proxy counterparts. The proxy--text grid contains 257 cases, including 60 reference cases. We add 15 no-image, 8 no-audio, and 18 no-proxy cases, plus 14 targeted cases that vary joint reference removal, identity, voice, event timing, or camera control. We separately analyze the Studio 3D proxy subset across five models, while the targeted input variations are examined individually in the appendix. The configurations allow us to compare the effects of proxy availability, input detail, and reference grounding. Full case assignments and evaluation protocols are provided in the appendix.

\subsection{Evaluation Framework}

\begin{figure*}[t]
    \centering
    \includegraphics[width=\linewidth]{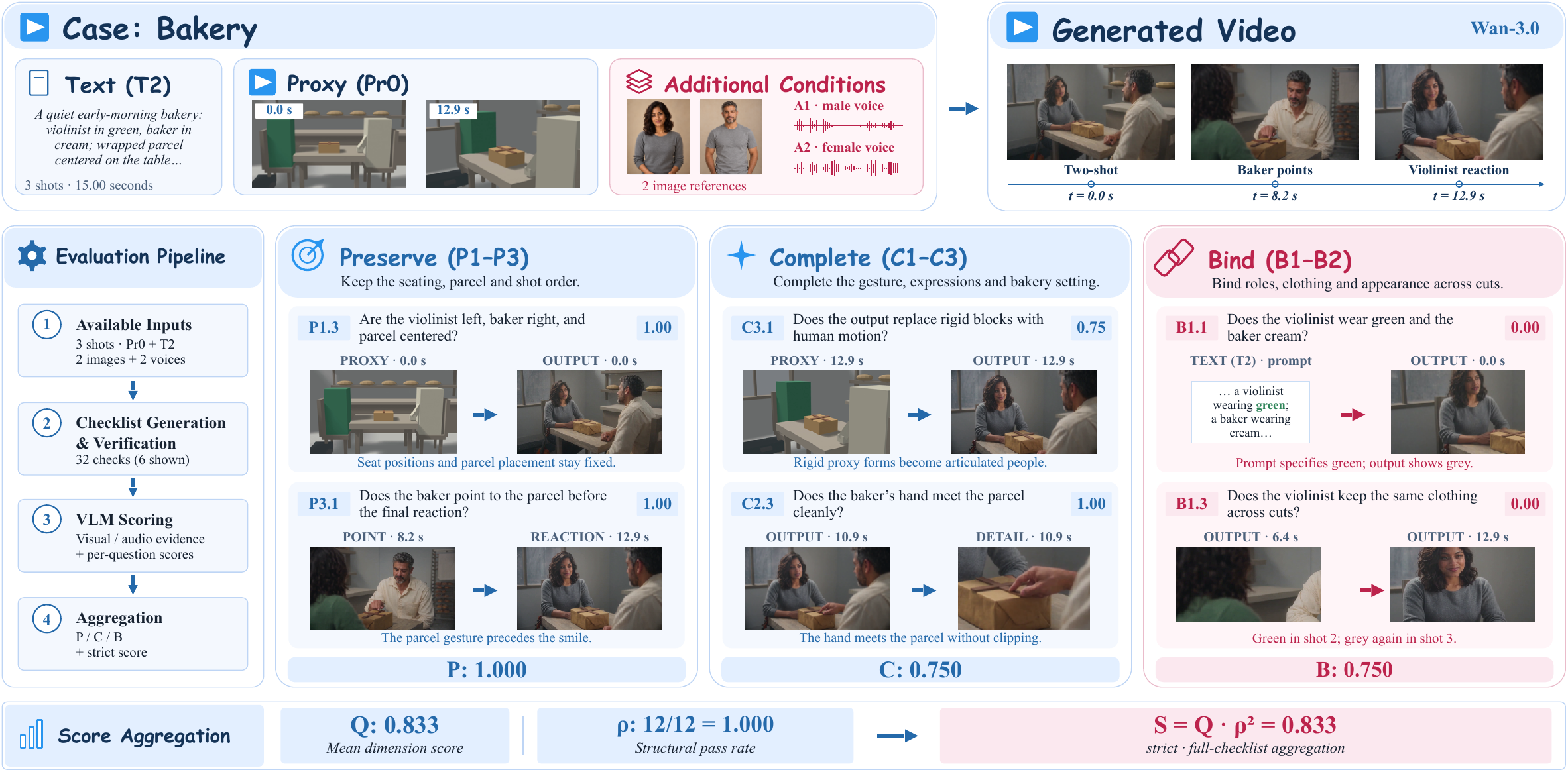}
    \caption{
    \textbf{Checklist-guided evaluation on the bakery case.} A case-specific checklist is generated and verified from the proxy, text, two image references, and two voice references, then used to score the Wan-3.0 video. Six of the 32 criterion-level assessments are illustrated with visual evidence, including the baker’s parcel gesture and the violinist’s clothing change across cuts. The full checklist yields \(Q=0.833\). All 12 critical Preserve criteria pass, giving \(\rho=12/12=1\) and Strict \(S=Q\rho^2=0.833\).
}
    \label{fig:evaluation}
\end{figure*}

A proxy specifies only part of the intended video, so matching the proxy alone does not guarantee successful generation. Instead, we use case-specific checklists to assess whether the generated video retains the intended structure, completes or corrects unfinished content, and binds supplementary conditions to the correct entities and events (Table~\ref{tab:eval_dimensions} and Fig.~\ref{fig:evaluation}).

\begin{table*}[t]
\centering
\small
\setlength{\tabcolsep}{5pt}
\caption{\textbf{Evaluation structure of ProxyGenBench.} Three dimensions comprise eight capabilities and 32 fine-grained criteria, with four criteria per capability. The table summarizes each capability's evaluation focus; criterion definitions are provided in the appendix. Case-specific checklist questions instantiate the applicable criteria.}
\label{tab:eval_dimensions}
\begin{tabularx}{\textwidth}{@{}lllX@{}}
\toprule
\textbf{Dimension} & \textbf{ID} & \textbf{Capability} & \textbf{Evaluation focus} \\
\midrule
Preserve & P1 & Spatial Structure & Proxy-defined scene layout, relative object positions, and spatial anchors. \\
 & P2 & Motion \& Camera Control & Proxy-defined coarse object trajectories, actions, and camera behavior. \\
 & P3 & Temporal \& Shot Continuity & Proxy-defined event order and object-state continuity across transitions. \\
\midrule
Complete & C1 & Visual Realization Quality & Plausible appearance, geometry, texture, and visual coherence. \\
 & C2 & Motion \& Physical Plausibility & Natural fine-grained articulation, physical interactions, and contact consistency. \\
 & C3 & Proxy Abstraction Handling & Completion of proxy placeholders and correction of proxy-induced artifacts. \\
\midrule
Bind & B1 & Entity \& Appearance Grounding & Correct assignment of textual and visual attributes to entities. \\
 & B2 & Action \& Audio-event Grounding & Correct assignment of actions, voices, and sound events to entities and moments. \\
\bottomrule
\end{tabularx}
\vspace{-0.5cm}
\end{table*}

\paragraph{Evaluation dimensions.} \textbf{Preserve} (P1--P3) tracks whether intended scene structure, motion, camera behavior, and event continuity survive generation. \textbf{Complete} (C1--C3) assesses whether the incomplete proxy is realized as coherent appearance, natural articulation, and plausible interactions instead of carrying over abstraction artifacts. \textbf{Bind} (B1--B2) assesses whether text, image, and audio conditions are correctly associated with entities, actions, and moments within that same structure. In the bakery example (Fig.~\ref{fig:evaluation}), the model must preserve the seating arrangement and event order, turn rigid proxy gestures into natural interactions, and bind the specified clothing and voice references to the correct characters. The baker may point to the parcel at the right moment while the violinist still appears in the wrong outfit. This shows how preserving the intended structure can coexist with incomplete realization or incorrect condition binding.

\paragraph{Checklist construction.} Eight capabilities comprise 32 fine-grained criteria: 12 for \textit{Preserve}, 12 for \textit{Complete}, and 8 for \textit{Bind}. For each case, Gemini-3.1-Pro~\cite{gemini31pro} instantiates applicable criteria into input-grounded questions~\cite{tang2026rewardverse, tang2026evolutionary, bai2026think}, potentially with multiple subquestions about entities, events, or intervals. Qwen3.5-397B-A17B~\cite{qwen35} verifies clarity, support from the available inputs, and alignment with the intended criterion; invalid questions are revised, with human feedback used to refine the procedure. Checklists are fixed before video scoring and shared across models on the same case. Each applicable criterion contributes one assessment.

\paragraph{Video scoring.} The Gemini judge scores each criterion based on visual or audio evidence. Directly verifiable requirements receive binary scores, while appearance and motion quality are evaluated using predefined grading scales. In Fig.~\ref{fig:evaluation}, the rigid-motion correction receives a graded score of 0.75 even though the event-order check passes, illustrating why proxy adherence alone does not determine completion quality. Reference-dependent criteria compare generated appearance, voices, or sound events with the supplied inputs. Inapplicable criteria are excluded; scoring prompts and rubrics are provided in the appendix.

\begin{figure*}
    \centering
    \includegraphics[width=0.83\linewidth]{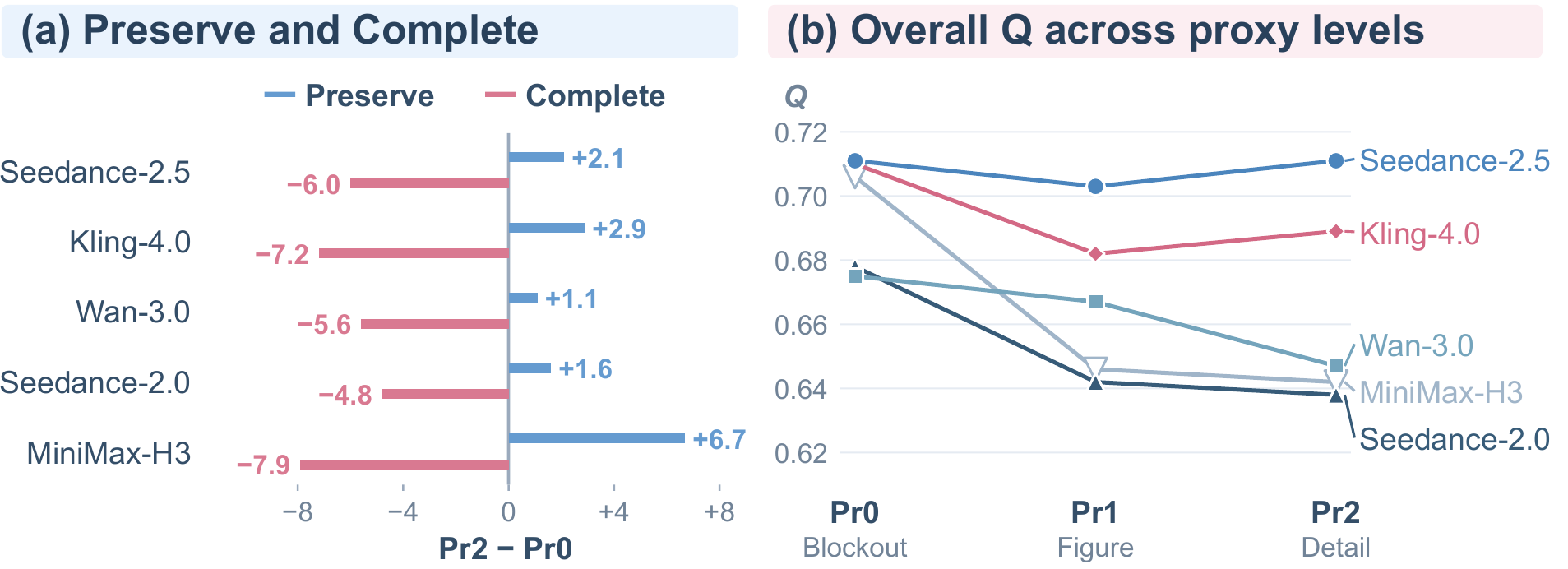}  
    \caption{\textbf{Proxy-fidelity effects on shared T2 scenes.} (a) Preserve and Complete changes from Pr0 to Pr2, in percentage points. (b) Mean dimension score $Q$ across proxy levels. The full $P/C/B/Q$ results appear in Table~\ref{tab:proxy_fidelity}.}
    \label{fig:proxy_fidelity_tradeoff}
    \vspace{-0.5cm}
\end{figure*}

\subsection{Score Aggregation}

\textit{Preserve}, \textit{Complete}, and \textit{Bind} provide separate diagnostic scores. We further combine them into a structure-aware score, since even a visually realistic video may fail to meet the creator's intent if it breaks the intended scene layout or event order. We therefore introduce the \textit{Strict} score, which places greater emphasis on critical \textit{Preserve} requirements while retaining the three separate dimension scores.

\paragraph{Hierarchical aggregation.} Each applicable criterion contributes one score, regardless of the number of subquestions. For case $i$, let $s_{i,k}\in[0,1]$ denote the score for criterion $k$. Using equal weights $a_k=1$, applicable criteria are aggregated into capability scores:

\begin{equation}
\qquad f_{i,g}=\frac{\sum_{k\in\mathcal{K}_{i,g}}a_k s_{i,k}}{\sum_{k\in\mathcal{K}_{i,g}}a_k}.
\end{equation}
Here, $\mathcal{K}_{i,g}$ contains the criteria applicable to capability $g$; inapplicable criteria are omitted. The resulting capability scores form \textit{Preserve}, \textit{Complete}, and \textit{Bind} dimension scores $D_{i,d}$ using the fixed aggregation specified in the appendix.

\begin{table}[t]
\centering
\caption{\textbf{Effect of proxy guidance.} (a) Control metrics on overlapping subsets (Pairs/Sc.: pairs/scenes; $\Delta$: proxy minus no proxy; scores in $[0,1]$). (b) VLM preferences over 50 pairs; Other includes ties and order-dependent outcomes.}
\label{tab:proxy_benefits}
\begingroup
\footnotesize
\setlength{\tabcolsep}{3pt}
\renewcommand{\arraystretch}{1.16}
\textbf{(a) Spatiotemporal control}\par\vspace{3pt}
\begin{tabular*}{\columnwidth}{@{\extracolsep{\fill}}lrrrr@{}}
\toprule[1.05pt]
\textbf{Metric} & \textbf{Pairs/Sc.} & \textbf{No proxy} & \textbf{+ Proxy} & \boldmath$\boldsymbol{\Delta}$ \\
\midrule[0.60pt]
Action transition $\uparrow$ & 22/8 & 0.894 & \textbf{1.000} & +0.106 \\
Cross-shot spatial $\uparrow$ & 14/5 & 0.833 & \textbf{0.952} & +0.119 \\
MEt3R $\downarrow$ & 17/10 & 0.464 & \textbf{0.337} & $-0.127$ \\
\bottomrule[1.05pt]
\end{tabular*}
\par\vspace{6pt}
\textbf{(b) VLM preference}\par\vspace{3pt}
\begin{tabular*}{\columnwidth}{@{\extracolsep{\fill}}lrrr@{}}
\toprule[1.05pt]
\textbf{Protocol} & \textbf{Proxy win} & \textbf{Other} & \textbf{Proxy loss} \\
\midrule[0.60pt]
Primary order & 40 & 2 & 8 \\
Two-order aggregate & \textbf{35} & 15 & 0 \\
\bottomrule[1.05pt]
\end{tabular*}
\endgroup
\vspace{-0.5cm}
\end{table}

\paragraph{Strict Score and Joint Success Rate.} We use the 12 predefined \textit{Preserve} criteria (P1.1--P3.4) as critical structural checks, with pass thresholds set before evaluation. For each case $i$, let $\mathcal{D}_i$ be the set of applicable dimensions and $\rho_i$ be the fraction of the 12 critical \textit{Preserve} checks that pass. We define the \textit{Average} score $Q_i$ and \textit{Strict} score $S_i$ as follows:

\begin{equation}
Q_i = \frac{1}{|\mathcal{D}_i|}
\sum_{d \in \mathcal{D}_i} D_{i,d},
\qquad
S_i = Q_i \rho_i^2.
\end{equation}
The factor $\rho_i^2$ gives greater weight to structural preservation: failing multiple critical checks reduces the overall score, even if the video looks complete or satisfies other conditions. Thus, \textit{Strict} emphasizes proxy control while keeping \textit{Preserve}, \textit{Complete}, and \textit{Bind} as separate diagnostic scores. We discuss this weighting and alternatives in the appendix. \textit{Joint Success} rate measures the fraction of outputs that pass all applicable critical \textit{Preserve} checks ($\rho_i=1$), rather than all P/C/B requirements. We compute model-level \textit{Strict} by averaging case scores within each input group. Missing model outputs receive a score of zero, while cases that cannot be evaluated are excluded.

\begin{table}[t]
\centering
\caption{\textbf{Overall proxy-conditioned performance.} $P/C/B$ denote Preserve/Complete/Bind; Average $Q$ is the mean of applicable dimensions and Strict $=Q\times(\text{critical pass rate})^2$. Joint is the fraction passing all 12 critical Preserve criteria. Missing model outputs receive zero. Scores lie in $[0,1]$ (higher is better); bold marks the best score within each group.}
\label{tab:overall}
\begingroup
\scriptsize
\setlength{\tabcolsep}{2.2pt}
\renewcommand{\arraystretch}{1.20}
\begin{tabular*}{\columnwidth}{@{\extracolsep{\fill}}lrrrrrr@{}}
\toprule[1.05pt]
\textbf{Model} & \textbf{$P$} & \textbf{$C$} & \textbf{$B$} & \textbf{Average} & \textbf{Strict} & \textbf{Joint} \\
\midrule[0.60pt]
\multicolumn{7}{@{}l}{\textbf{(a) Without independent audio}} \\
\addlinespace[1pt]
Seedance-2.5 & \textbf{0.774} & 0.566 & \textbf{0.693} & \textbf{0.678} & \textbf{0.450} & 0.140 \\
Kling-4.0 & 0.746 & \textbf{0.591} & 0.652 & 0.663 & 0.425 & 0.150 \\
Wan-3.0 & 0.732 & 0.557 & 0.642 & 0.644 & 0.399 & 0.164 \\
Seedance-2.0 & 0.698 & 0.505 & 0.642 & 0.615 & 0.375 & \textbf{0.174} \\
MiniMax-H3 & 0.639 & 0.539 & 0.667 & 0.615 & 0.344 & 0.072 \\
\midrule[0.60pt]
\multicolumn{7}{@{}l}{\textbf{(b) With independent audio}} \\
\addlinespace[1pt]
Seedance-2.5 & \textbf{0.645} & \textbf{0.612} & \textbf{0.740} & \textbf{0.666} & \textbf{0.325} & 0.040 \\
Seedance-2.0 & 0.633 & 0.550 & 0.718 & 0.634 & 0.314 & \textbf{0.080} \\
Wan-3.0 & 0.617 & 0.511 & 0.624 & 0.584 & 0.324 & \textbf{0.080} \\
MiniMax-H3 & 0.553 & 0.536 & 0.654 & 0.581 & 0.241 & 0.020 \\
\bottomrule[1.05pt]
\end{tabular*}
\endgroup
\vspace{-0.5cm}
\end{table}

\section{Experiments}
We first assess the control benefits of structural proxies, then examine joint structural constraints, proxy fidelity, text detail, and condition binding. We finally compare the VLM evaluator with human judgments.

\begin{table*}[t]
\centering
\caption{\textbf{Results across three proxy-fidelity levels.} $P/C/B$ denote Preserve/Complete/Bind, and \textbf{Average} is their mean. For each model and metric, we \textbf{bold} the best score across fidelity levels to highlight which level performs best for each metric.}
\label{tab:proxy_fidelity}
\begingroup
\scriptsize
\setlength{\tabcolsep}{2.4pt}
\renewcommand{\arraystretch}{1.18}
\begin{tabular*}{\textwidth}{@{\extracolsep{\fill}}l*{12}{r}@{}}
\toprule[1.10pt]
& \multicolumn{4}{c}{\textbf{Pr0 -- Blockout}}
& \multicolumn{4}{c}{\textbf{Pr1 -- Figure}}
& \multicolumn{4}{c}{\textbf{Pr2 -- Detail}} \\
\cmidrule[0.48pt](lr){2-5}
\cmidrule[0.48pt](lr){6-9}
\cmidrule[0.48pt](lr){10-13}
\textbf{Model}
& \textbf{$P$} & \textbf{$C$} & \textbf{$B$} & {Average}
& \textbf{$P$} & \textbf{$C$} & \textbf{$B$} & {Average}
& \textbf{$P$} & \textbf{$C$} & \textbf{$B$} & {Average} \\
\midrule[0.60pt]
Seedance-2.5
& 0.780 & \textbf{0.598} & 0.754 & \textbf{0.711}
& \textbf{0.801} & 0.597 & 0.714 & 0.704
& \textbf{0.801} & 0.538 & \textbf{0.794} & \textbf{0.711} \\
Kling-4.0
& 0.753 & \textbf{0.647} & \textbf{0.730} & \textbf{0.710}
& 0.774 & 0.610 & 0.661 & 0.682
& \textbf{0.782} & 0.575 & 0.710 & 0.689 \\
Wan-3.0
& 0.755 & \textbf{0.587} & \textbf{0.681} & \textbf{0.675}
& \textbf{0.782} & 0.553 & 0.665 & 0.667
& 0.766 & 0.531 & 0.645 & 0.647 \\
Seedance-2.0
& 0.737 & \textbf{0.560} & \textbf{0.738} & \textbf{0.678}
& \textbf{0.788} & 0.482 & 0.657 & 0.642
& 0.753 & 0.512 & 0.649 & 0.638 \\
MiniMax-H3
& 0.664 & \textbf{0.612} & \textbf{0.843} & \textbf{0.706}
& \textbf{0.737} & 0.528 & 0.673 & 0.646
& 0.731 & 0.533 & 0.661 & 0.642 \\
\bottomrule[1.10pt]
\end{tabular*}
\endgroup
\vspace{-0.3cm}
\end{table*}

\begin{table*}[t]
\centering
\caption{\textbf{Strict scores across input configurations.} (a) Text-detail levels at Pr0 without independent audio references; bold indicates the highest score for each model. (b) Multimodal configurations, where Pr, T, I, and A denote proxy, text, image references, and independent audio references. The +I, +A, and +I+A columns add the indicated references to Pr+T. '-' denotes an unsupported input.}
\label{tab:input_conditions}
\begingroup
\small
\setlength{\tabcolsep}{5pt}
\renewcommand{\arraystretch}{1.20}
\begin{tabular*}{\textwidth}{@{\extracolsep{\fill}}lrrr@{\hspace{17pt}}lrrrr@{}}
\toprule[1.10pt]
\multicolumn{4}{c}{\textbf{(a) Text detail}} & \multicolumn{5}{c}{\textbf{(b) Multimodal inputs}} \\
\cmidrule[0.48pt](lr){1-4}\cmidrule[0.48pt](lr){5-9}
\textbf{Model} & \textbf{T0} & \textbf{T1} & \textbf{T2} & \textbf{Model} & \textbf{Pr+T} & \textbf{+I} & \textbf{+A} & \textbf{+I+A} \\
\midrule[0.60pt]
Seedance-2.5 & 0.434 & 0.454 & \textbf{0.458} & Seedance-2.5 & \textbf{0.454} & 0.391 & 0.365 & 0.256 \\
Kling-4.0 & 0.394 & 0.426 & \textbf{0.437} & Kling-4.0 & \textbf{0.449} & 0.347 & -- & -- \\
Wan-3.0 & 0.348 & 0.382 & \textbf{0.425} & Wan-3.0 & \textbf{0.438} & 0.352 & 0.349 & 0.394 \\
Seedance-2.0 & 0.325 & \textbf{0.385} & 0.375 & Seedance-2.0 & \textbf{0.431} & 0.265 & 0.199 & 0.191 \\
MiniMax-H3 & 0.307 & 0.336 & \textbf{0.375} & MiniMax-H3 & \textbf{0.392} & 0.290 & 0.252 & 0.266 \\
\bottomrule[1.10pt]
\end{tabular*}
\endgroup
\vspace{-0.5cm}
\end{table*}

\subsection{Experimental Setup}
\paragraph{Models and protocol.} We evaluate Seedance-2.5~\cite{bytedance2026seedance25}, Kling-4.0~\cite{kuaishou2026kling4}, Wan-3.0~\cite{alibaba2026wan3}, Seedance-2.0~\cite{bytedance2026seedance20}, and MiniMax-H3~\cite{minimax2026h3} on supported configurations among the 294 proxy-conditioned cases. Results are grouped by independent audio-reference availability; Kling-4.0 is excluded from the audio-reference group. Models within each group share cases and verified checklists. \textit{Strict} provides the primary structure-sensitive ranking, while the 18 no-proxy cases are used for matched comparisons. Missing model outputs receive zero, as specified in Sec.~3.4. Model settings and evaluation details are provided in the appendix.

\paragraph{Proxy availability.} We evaluate 50 model-level matched proxy/no-proxy video pairs from 18 scenes, holding text, other references, and generation settings fixed. Action-transition, spatial-consistency, and geometric metrics use their applicable, potentially overlapping subsets. VLM preferences are aggregated across both presentation orders to reduce order bias.

\paragraph{Cross-condition comparisons.} Adding text detail or references can change the requirements applicable to a case, especially for Bind. We therefore compare matched configurations on shared checklist criteria and assess newly introduced binding requirements separately. Scores from different case sets or checklists are not interpreted as isolated modality effects. Details are in the appendix.

\subsection{Proxy Benefits}
The structural prior helps when prompts leave temporal and spatial relationships ambiguous (Table~\ref{tab:proxy_benefits}). On matched subsets, action-transition accuracy rises from 0.894 to 1.000, cross-shot spatial consistency from 0.833 to 0.952, and MEt3R~\cite{asim25met3r} error falls from 0.464 to 0.337. Under the two-order VLM protocol, proxy-guided outputs are preferred in 35 of 50 pairs (70\%); the other outcomes are ties or order-dependent. These comparisons establish benefits for the measured structural controls, while motion quality and condition binding require separate evaluation.

\subsection{Joint Structural Success}
Average \textit{Preserve} scores do not show whether all critical structural checks hold in the same output (Table~\ref{tab:overall}). Without independent audio references, Seedance-2.5 and Kling-4.0 lead \textit{Strict} (0.450 and 0.425), yet their \textit{Joint Success} rates are only 14.0\% and 15.0\%. Their strengths also differ: Seedance-2.5 leads Preserve and Bind, whereas Kling-4.0 leads Complete. Across all five models, \textit{Joint Success} ranges from 7.2\% to 17.4\%, with Seedance-2.0 highest despite a lower \textit{Strict} score. This ranking reversal reveals a gap between structure-sensitive overall performance and joint constraint satisfaction: a model can achieve a higher \textit{Strict} score without satisfying all critical structural constraints together.

\subsection{Proxy Fidelity}
More detailed proxies do not consistently improve the final video (Table~\ref{tab:proxy_fidelity} and Fig.~\ref{fig:proxy_fidelity_tradeoff}). On the 31 shared T2 scenes without independent audio references, \textit{Preserve} increases by 1.1--6.7 percentage points from Pr0 to Pr2 across all five models, while \textit{Complete} decreases by 4.8--7.9 points.
This tension persists even in the leading models: Seedance-2.5 gains 2.1 points in \textit{Preserve} but loses 6.0 in \textit{Complete}, while Kling-4.0 gains 2.9 but loses 7.2. Their overall responses differ, however. Seedance-2.5 maintains an Average score of 0.711 as \textit{Bind} improves from 0.754 to 0.794, whereas Kling-4.0 drops from 0.710 to 0.689. Pr0 yields the highest \textit{Average} for the other four models. Thus, stronger overall performance does not eliminate the preservation–completion tension, but models differ in how they balance these requirements.

\begin{table}[t]
\centering
\vspace{-0.2cm}
\caption{\textbf{Performance on Studio 3D proxy cases.} $P$, $C$, and $B$ denote Preserve, Complete, and Bind; Strict $=Q\times\text{critical pass rate}^2$. Scores lie in $[0,1]$ (higher is better); bold marks the highest score in each column.}
\label{tab:targeted_cases}
\small
\setlength{\tabcolsep}{4pt}
\renewcommand{\arraystretch}{1.08}
\begin{tabular*}{\linewidth}{@{\extracolsep{\fill}}lrrrr@{}}
\toprule
\textbf{Model} & $P$ & $C$ & $B$ & Strict \\
\midrule
Seedance-2.5 & 0.924 & 0.511 & 0.318 & 0.505 \\
Kling-4.0 & 0.894 & 0.633 & 0.318 & 0.525 \\
Wan-3.0 & 0.947 & \textbf{0.665} & 0.273 & \textbf{0.564} \\
Seedance-2.0 & \textbf{0.962} & 0.477 & 0.182 & 0.498 \\
MiniMax-H3 & 0.848 & 0.616 & \textbf{0.500} & 0.536 \\
\bottomrule
\end{tabular*}
\vspace{-0.5cm}
\end{table}

More detailed proxies specify useful structure but may also contain rigid poses, awkward contacts, or unnatural motion. Following those local details too closely can carry proxy artifacts into the final video instead of correcting them. These opposing trends highlight a central challenge: generators must follow scene-defining constraints without copying unintended proxy details.

\subsection{Text Detail}
Text supplies semantic information that proxy geometry and motion leave unspecified (Table~\ref{tab:input_conditions}(a)). Under Pr0 without independent audio references, T2 scores higher than T0 on \textit{Strict} for all five models, although Seedance-2.0 peaks at T1. The observed T0-to-T2 gains are larger for Wan-3.0 (+7.7 points) and MiniMax-H3 (+6.8) than for Seedance-2.5 (+2.4), narrowing their gaps to the leading model. These gains suggest that detailed text can compensate for weaker baselines, rather than benefiting all models equally.

\subsection{Condition Binding}
Following the proxy does not guarantee that supplementary conditions are assigned to the intended entities and events.

\paragraph{Image references.} On their applicable subsets, identity matching and cross-shot identity consistency average 0.897 and 0.909, whereas action--object binding averages only 0.636. Some generated videos swap face identities while retaining the original clothing assignments. Thus, recognizing reference attributes is distinct from binding those attributes to the right entities and actions. All five models also obtain lower observed Strict scores on Pr+T+I cases than on Pr+T cases (Table~\ref{tab:input_conditions}(b)), reinforcing the difficulty of integrating visual references into an already structured scene (see the appendix for detailed analyses).

\paragraph{Audio references.} Voice/source matching averages 0.768, while audiovisual timing alignment reaches only 0.513. A model may reproduce the intended voice without placing it at the correct moment. This reveals a temporal binding failure that structural guidance alone does not resolve.

\paragraph{Studio 3D proxies.}
\label{sec:targeted_cases}
On the 12 Studio 3D proxy cases evaluated across all five models (Table~\ref{tab:targeted_cases}), \textit{Preserve} ranges from 0.848 to 0.962, whereas \textit{Bind} reaches only 0.182--0.500. Seedance-2.5 and Kling-4.0, the overall Strict leaders without independent audio, both score $B=0.318$, below MiniMax-H3 ($B=0.500$). Seedance-2.0 attains $P=0.962$ but only $B=0.182$. Proxy fidelity reveals a similar separation: \textit{Preserve} rises from Pr0 to Pr2, yet \textit{Bind} drops by 3.6 points for Wan-3.0 and 18.2 for MiniMax-H3 (Table~\ref{tab:proxy_fidelity}). Together, these results show that condition binding is not a by-product of structural control: models must ground supplementary information within the structure they preserve.

\begin{table}[t]
\centering
\vspace{-0.2cm}
\caption{\textbf{Human validation on 3,199 judgments from 110 videos.} Judg. counts audited judgments; Valid and Acc. denote checklist validity and scoring accuracy against human annotations. Pass and Fail report accuracy within judge-labeled pass and fail groups.}
\label{tab:human_validation}
\begingroup
\footnotesize
\setlength{\tabcolsep}{3.4pt}
\renewcommand{\arraystretch}{1.18}
\begin{tabular*}{\columnwidth}{@{\extracolsep{\fill}}lrrrrr@{}}
\toprule[1.05pt]
\textbf{Dimension} & \textbf{Judg.} & \textbf{Valid} & \textbf{Acc.} & \textbf{Pass} & \textbf{Fail} \\
\midrule[0.60pt]
Preserve & 1,320 & 0.998 & 0.897 & 0.902 & 0.865 \\
Complete & 1,320 & 1.000 & 0.952 & 0.963 & 0.916 \\
Bind & 559 & 1.000 & 0.948 & 0.981 & 0.840 \\
\midrule[0.60pt]
\textbf{Overall} & \textbf{3,199} & \textbf{0.999} & \textbf{0.929} & \textbf{0.939} & \textbf{0.885} \\
\bottomrule[1.05pt]
\end{tabular*}
\endgroup
\vspace{-0.5cm}
\end{table}

\subsection{Human Validation}
We audit 3,199 criterion-level judgments from 110 videos (Table~\ref{tab:human_validation}). Checklist validity is 99.9\%, and overall scoring accuracy against human annotations reaches 92.9\%. Scoring accuracy is lowest for Preserve (89.7\%); for Bind, accuracy on judge-labeled failures is 84.0\%, compared with 98.1\% for passes. This gap matters because the benchmark must detect subtle structural violations and incorrect grounding. We therefore report dimension-level and pass/fail accuracy rather than relying only on an aggregate score; further annotation details are provided in the appendix.
\section{Conclusion}
ProxyGenBench evaluates three interacting requirements of video generation from imperfect structural proxies: Preserve, Complete, and Bind. Across five models, proxy guidance improves spatiotemporal control, yet strong average adherence can mask violations of joint structural constraints. More detailed proxies can also reduce completion scores, while preserved structure does not ensure correct condition binding. These findings highlight the need for selective control: generators must retain intended structural constraints while correcting proxy artifacts and grounding supplementary information in the same scene.

{
    \small
    \bibliographystyle{ieeenat_fullname}
    \bibliography{main}
}

\appendix
\clearpage
\maketitlesupplementary

\begin{figure*}
    \centering
    \vspace{-0.2cm}
    \includegraphics[width=\linewidth]{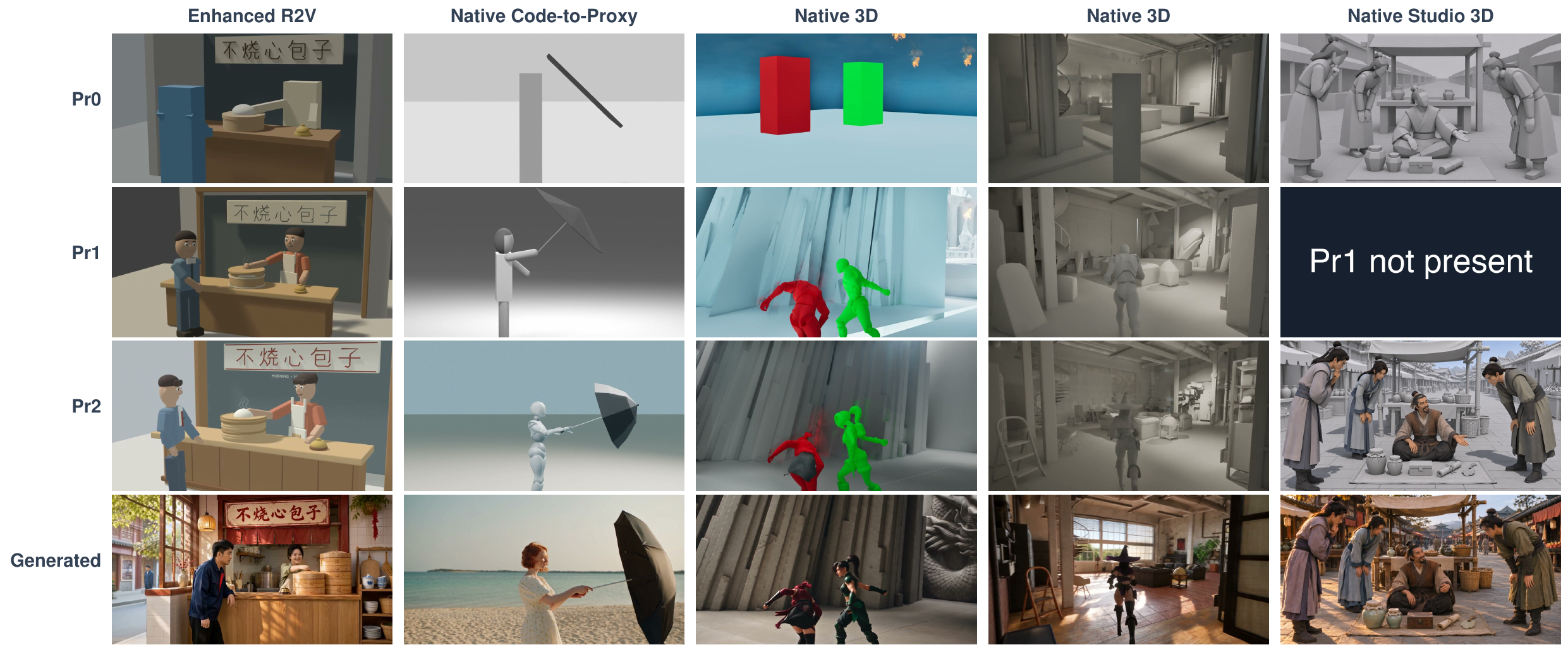}
    \caption{Representative proxy fidelity levels and corresponding Seedance-2.5 outputs. The Studio 3D example has no $Pr_1$ proxy.}
    \label{fig:proxy_level_examples}
    \vspace{-0.3cm}
\end{figure*}

\section{Benchmark Construction}
\subsection{Proxy and Text Fidelity}

\paragraph{Proxy levels.}
For each scene, we define up to three proxy levels, \(Pr_0\), \(Pr_1\), and \(Pr_2\). When multiple levels are available, they preserve the same story, shot structure, camera plan, blocking, role colours, and key trajectories, while varying primarily in geometric detail and motion fidelity. For Blender-authored proxies, \(Pr_0\) uses coarse coloured primitives; \(Pr_1\) adds articulated bodies and readable props; and \(Pr_2\) remains untextured but provides clearer geometry and more faithful motion. For proxies from other sources, the available levels may have source-specific representations and may not span all three levels; nevertheless, they follow the same abstract-to-detailed ordering, with \(Pr_0\) denoting the most abstract representation, as shown in \cref{fig:proxy_level_examples}.

\paragraph{Text prompt levels.}
We define three text-prompt levels, \(T_0\), \(T_1\), and \(T_2\), which progressively increase the amount of semantic and temporal information supplied beyond the proxy video (Figure~\ref{fig:prompt-levels}). \(T_0\) provides a concise target description---including the intended visual domain, principal subjects, setting, and key transformation---while relying primarily on the proxy for shot order, camera framing, blocking, and motion. \(T_1\) additionally specifies the desired appearance of characters, objects, and environment, together with coarse shot-level actions and temporal beats. \(T_2\) gives the most complete specification: detailed visual attributes, physical and interaction constraints, cinematographic instructions, negative constraints, audio requirements when applicable, and a time-resolved plan aligned with the proxy. Across all levels, the proxy remains the primary structural signal; the prompt directs the model to recreate the intended scene with the desired appearance and realism.

\paragraph{Controlled case construction.}
Each scene is organized as a controlled experiment with a shared target story, duration, generation settings, and reference conditions. The possible core configurations (when the corresponding proxy levels are available) are
\[
\{Pr_0\text{-}T_0,\;Pr_0\text{-}T_1,\;Pr_0\text{-}T_2,\;Pr_1\text{-}T_2,\;Pr_2\text{-}T_2\}.
\]
The \(Pr_0\text{-}T_{0,1,2}\) cases vary text specificity at fixed proxy fidelity, while \(Pr_{0,1,2}\text{-}T_2\) varies proxy detail and motion fidelity at fixed detailed text.

In addition to the proxy--text grid and the input ablations described below, we include 14 targeted control cases. Each keeps the same scene goal but changes one specific input or instruction. Four cases test voice assignment in two dialogue scenes: in the normal version, each character uses their designated voice reference; in the swapped version, the two characters exchange voice references. The remaining cases test whether models follow a specific identity reference, assign the correct image to the correct role, delay the closing of a gate, distinguish camera movement from zoom, shift focus at the intended time, keep a centered camera while moving through a doorway, or operate without both image and audio references. These controls test whether a model uses each condition for its intended purpose rather than relying on an easier but incorrect cue.

\begin{table*}[t]
\centering
\caption{Benchmark configurations and controlled comparisons. Text and proxy level counts follow the order $T_0/T_1/T_2$ and $Pr_0/Pr_1/Pr_2$, respectively. The fidelity analysis uses 31 matched scenes.}
\label{tab:supp_case_grid}
\footnotesize
\setlength{\tabcolsep}{6pt}
\renewcommand{\arraystretch}{1.18}
\begin{tabularx}{\textwidth}{@{}
>{\raggedright\arraybackslash}p{.18\textwidth}
>{\raggedright\arraybackslash}p{.19\textwidth}
>{\raggedright\arraybackslash}p{.23\textwidth}
>{\raggedright\arraybackslash}X@{}}
\toprule
\textbf{Contrast} & \textbf{Cases} & \textbf{Fixed conditions} & \textbf{Research question} \\
\midrule
Text detail & $Pr_0$: 50/50/60 & Proxy fidelity, scene intention & Does more explicit text improve realization of the proxy-defined sequence? \\
\addlinespace[3pt]
Proxy detail & $T_2$: 60/47/50 & Target text, scene, references & How does added geometric and motion detail affect preservation and completion? \\
\addlinespace[3pt]
Reference removal & 15 no-image; 8 no-audio & Scene and remaining conditions & Which identity, appearance, voice, and event requirements depend on references? \\
\addlinespace[3pt]
Proxy removal & 18 no-proxy & Target description, remaining references, generation settings & How much does structural guidance help motion, camera, and continuity? \\
\addlinespace[3pt]
Targeted controls & 14 variations & Base scene apart from the targeted change & Are identity, voice, event timing, and camera instructions grounded correctly? \\
\bottomrule
\end{tabularx}
\vspace{-0.5cm}
\end{table*}

\definecolor{PromptBlue}{HTML}{1F4E79}
\definecolor{PromptTeal}{HTML}{0E7490}
\definecolor{PromptViolet}{HTML}{6D28D9}
\definecolor{PromptInk}{HTML}{1F2937}

\begin{figure*}[!t]
  \centering
  \setlength{\tabcolsep}{0pt}
  \begin{minipage}{0.985\textwidth}
    \small
    \textbf{Representative prompt-level progression.}
    All examples below use the same low-detail proxy \(Pr_0\), the same umbrella reference image, and the same 15-second target. They illustrate how increasing text specificity complements---rather than replaces---the proxy's shot and motion signal.
    
    \vspace{0.35em}
    \begin{tcolorbox}[
      enhanced, breakable, colback=PromptBlue!3, colframe=PromptBlue,
      boxrule=0.7pt, arc=2pt, left=7pt, right=7pt, top=5pt, bottom=5pt,
      title=\textbf{\(T_0\) --- concise scene directive},
      coltitle=white, colbacktitle=PromptBlue, fonttitle=\small
    ]
    \textit{Use the attached proxy video for shot order, timing, camera framing and rough body and umbrella motion; use reference image 1 for the canopy's black colour, eight-rib geometry, curvature and glossy finish. Replace the proxy figures and background with natural live action.}

    Create a 15-second warm live-action beach advertisement. One young fair-skinned woman with short wavy auburn hair wears a white knee-length dress with small yellow lemon-floral prints and puff sleeves. She has one folding umbrella with a glossy black eight-rib canopy, silver-grey shaft and matte black handle. On a pale sandy beach beside a calm turquoise sea, she turns toward the camera, opens the umbrella, moves with it and brings it above and behind her. Follow the proxy for the changing views and gestures, rendering the umbrella mechanism and her movements naturally. Keep her appearance and the umbrella consistent. Include soft breeze, distant waves and an opening pop; no music.
    \end{tcolorbox}

    \vspace{0.35em}
    \begin{tcolorbox}[
      enhanced, breakable, colback=PromptTeal!3, colframe=PromptTeal,
      boxrule=0.7pt, arc=2pt, left=7pt, right=7pt, top=5pt, bottom=5pt,
      title=\textbf{\(T_1\) --- controlled appearance and coarse timeline},
      coltitle=white, colbacktitle=PromptTeal, fonttitle=\small
    ]
    \textit{Treat the attached proxy reference video as guidance for motion, timing, framing and camera placement, not final appearance. Preserve its four beats, cut order, body orientations and umbrella positions without tracing rigid block movements. Reference image 1 supplies the canopy's glossy black colour, eight-rib geometry, curvature and finish only; exclude its white studio background.}

    Generate a \(1280\times720\), 15-second polished, warmly graded live-action beach advertisement. Exactly one young woman stands alone with one folding umbrella. Keep her identity, costume and the umbrella's construction consistent across all cuts. Use pale beige sand, a calm blue-green sea, a thin distant coastline and a clear lightly hazed sky. Movement is subtle and human; sound is diegetic sea breeze, distant waves and an umbrella-opening pop; no music.

    \textbf{0.00--about 3.67:} medium view; she turns gradually from a rear-facing pose toward camera, with the umbrella out of view. \quad
    \textbf{About 3.67--7.50:} side view; she presents the folded umbrella, opens it, then raises it overhead. \quad
    \textbf{About 7.50--11.25:} wider nearly frontal view; she moves the open umbrella beside her and lifts it outward and upward. \quad
    \textbf{About 11.25--15.00:} closer view; she turns toward camera with the open umbrella behind her shoulder and settles into a front-facing hero composition.
    \end{tcolorbox}

    \vspace{0.35em}
    \begin{tcolorbox}[
      enhanced, breakable, colback=PromptViolet!3, colframe=PromptViolet,
      boxrule=0.7pt, arc=2pt, left=7pt, right=7pt, top=5pt, bottom=5pt,
      title=\textbf{\(T_2\) --- full visual, physical, and temporal specification},
      coltitle=white, colbacktitle=PromptViolet, fonttitle=\small
    ]
    \textit{Use the attached proxy reference video only for motion, camera placement and timing. Preserve all four beats, cut order, framing, angles, body orientations and umbrella progression: out of view at first, then presented and opened, moved outward, and brought behind her shoulder. The proxy does not define appearance: replace its cuboid, slab, rigid motion, flat shading and two-colour backdrop with natural live action.}

    Create a \(1280\times720\), 25-fps, 15-second commercial-grade lifestyle beach advertisement. The woman has fair skin, short softly wavy auburn hair, a peaceful expression, and a white summer dress with yellow lemon-floral prints and short puff sleeves. Require natural skin, credible hands and limbs, quiet breathing, small weight shifts, and breeze-responsive hair and fabric. The folding umbrella begins compact and closed; its working slider and articulated, exactly eight ribs open once, and its glossy black fabric must retain curved, tensioned panels rather than becoming metallic or grey. Use warm natural light, soft coherent shadows, atmospheric depth, pale grainy sand, a turquoise blue-green sea, and a low distant coastline. Keep the woman alone; no unrelated activity, weather change, or music.

    \textbf{0.00--about 3.67:} medium view, rear-facing turn toward camera; umbrella absent. \quad
    \textbf{About 3.67--7.50:} side view, plausible slider-and-rib opening with natural hand contact and one soft pop. \quad
    \textbf{About 7.50--11.25:} wider nearly frontal view, the open umbrella travels beside her then outward and upward as her free arm extends. \quad
    \textbf{About 11.25--15.00:} closer rear-three-quarter-to-front turn, ending in a relaxed hero pose; canopy, eight ribs, face, and hands remain readable.
    \end{tcolorbox}
  \end{minipage}
  \caption{Text-detail levels for the \textsc{Umbrella Girl} case. The proxy and reference image remain fixed across $T_0$, $T_1$, and $T_2$.}
  \label{fig:prompt-levels}
  \vspace{-0.5cm}
\end{figure*}

\subsection{Proxy Construction}
\label{sec:supp_proxy_construction}
\paragraph{Source workflows.}
Proxies originate from four source groups. \emph{Enhanced R2V} adds animated structural guidance to reference-to-video tasks; \emph{Native Code-to-Proxy} animates three-dimensional scenes from scripts or optional reference clips through an LLM-based authoring process; \emph{Game Proxy} uses game footage or native game assets and animation; and \emph{Studio 3D Proxy} uses unfinished production scenes. These groups provide heterogeneous incompleteness and artifact patterns, rather than interchangeable samples of one authoring pipeline. The four workflows span reference-conditioned content, script-authored scenes, game environments, and unfinished animation assets, providing distinct sources of missing detail and imperfect motion.

\paragraph{Blender-authored proxies and native hierarchies.}
In the Blender-authored examples, lower-detail proxies contain color-coded primitives, while more detailed versions replace these with articulated figures, distinguishable props, and more informative local motion. Native 3D hierarchies and unfinished studio assets retain source-specific geometric representations and animation. When multiple levels are present, the target story and major camera and event structure are kept consistent; not every source provides every fidelity level. The level labels describe the amount of structural and motion detail, not the correctness of all depicted details.

\subsection{Scene Coverage}
\label{sec:supp_case_coverage}
Scene tasks identify the principal structural challenge: dynamic action, multi-shot storytelling, or multi-view consistency. Separately, five content-category annotations describe scene subject matter: Characters (22), Navigation (17), Objects (11), Combat \& Transform (7), and Camera (3). The proxy-source, task, and content-category annotations serve distinct roles and should not be interpreted as three mutually exclusive partitions of the same type.

The experimental contrasts are organized in Table~\ref{tab:supp_case_grid}. The 257 core cases are augmented with 18 no-proxy, 15 no-image, eight no-audio, and 14 targeted cases, giving 312 cases from 60 scenes. Of these, 294 include a proxy video and 18 omit it. Image and independent audio references appear in 177 and 60 cases, respectively; reference-type counts are non-exclusive. The Studio 3D subset contains 12 cases from 10 scenes and is analyzed separately in the main paper; these cases are part of the 312-case benchmark, rather than additional targeted variations. The case identifier, source, task, proxy and text levels, and optional reference indicators define the corresponding experimental configuration.

\subsection{Target Conditions}
\label{sec:supp_intent}
Text prompts communicate intended appearance, interactions, sound, and target style in addition to any structural requirements conveyed by the proxy. Reference images can specify identity, clothing, objects, and appearance, while independent audio references can specify voice/source identity or sound events. Such requirements are evaluated for the entity or event to which they are assigned, not merely for their appearance somewhere in the video.

\paragraph{Intended stylization versus artifacts.}
Proxy fidelity is not a style target by default. Intended stylization, including intentionally simplified geometry or stylized movement, is described in the accompanying text and is not counted as an unintended proxy artifact. By contrast, proxy placeholders, rigid pseudo-actions, clipping, or missing articulation are completion or correction targets only when the generator's supplied conditions support that expectation. A checklist must not introduce stylistic requirements from the hidden scene specification alone.

\paragraph{Condition roles.}
The T0--T2 umbrella progression in Fig.~\ref{fig:prompt-levels} is an actual
illustrative input from the benchmark. It exposes a useful separation between
\emph{where and when} (primarily controlled by the proxy), \emph{what must be
realized or repaired} (the target prompt), and \emph{which entity receives a
reference} (supplementary conditions). Table~\ref{tab:supp_prompt_contract}
records these distinct sources of evaluation evidence. In particular, a
low-poly look is not an error when stylization is explicitly requested; a
rigid placeholder is a completion target when natural articulation is
specified. This distinction is established before viewing model outputs.

\begin{figure*}
    \centering
    \includegraphics[width=\linewidth]{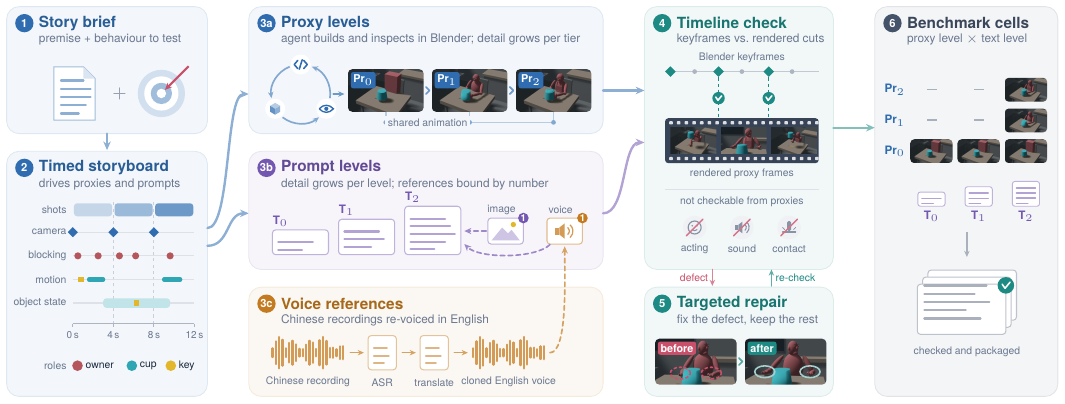}
    \caption{Proxy construction and repair pipeline, illustrated with \textit{The key under the cup}, a case without image or audio conditions. Optional reference branches show steps used for other applicable scenes. Checks and repairs are performed iteratively, with scene-specific evidence.}
    \label{fig:agent_construction}
\end{figure*}

\begin{table}[t]
\centering
\caption{Input conditions and evaluation roles.}
\label{tab:supp_prompt_contract}
\footnotesize
\setlength{\tabcolsep}{5pt}
\renewcommand{\arraystretch}{1.2}
\begin{tabularx}{\columnwidth}{@{}l>{\raggedright\arraybackslash}X l@{}}
\toprule
\textbf{Input} & \textbf{Intended constraint} & \textbf{Evaluation} \\
\midrule
\textbf{Proxy} & Layout, motion, camera, event order & Preserve \\
\textbf{Text} & Appearance, style, actions, corrections & Complete; Bind \\
\textbf{Image} & Identity and appearance of named targets & Bind (B1) \\
\textbf{Audio} & Voice/source, sound event, timing & Bind (B2) \\
\bottomrule
\end{tabularx}
\end{table}

\paragraph{Audio references.}\label{sec:supp_audio}
Audio-related checks distinguish source assignment (who speaks or produces a sound), event correctness (whether the specified sound occurs), and synchronization (whether the sound aligns with the intended visible event). Independent audio-reference configurations are reported separately from configurations without independent audio references, because input support and the applicable Bind requirements differ. Kling-4.0 is excluded from the audio-reference group because that input configuration is unsupported.

\subsection{Agent-Assisted Construction and Verification}
\label{sec:agent_construction}

Our construction process combines LLM-assisted authoring with iterative inspection and correction, as illustrated in \cref{fig:agent_construction}. An agent converts a shared scene specification into executable Blender scenes and English prompts. We then inspect the rendered proxies, source files, and packaged inputs, correcting inconsistencies before finalizing the benchmark cases. The figure illustrates the general workflow, although the inspection process and available evidence vary across scenes. Native 3D and Studio 3D sources use separate construction pipelines.

\paragraph{Shared scene specification.}
Construction begins with a story brief and the behavior to be tested, such as object persistence, role correspondence, or camera-motion control. The agent expands the brief into a timed storyboard specifying shot order, camera framing and movement, character blocking, key trajectories, and relevant object-state changes. This storyboard guides both proxy construction and prompt authoring (\cref{fig:agent_construction}, steps 1--2). Stable role colours identify characters and salient objects within the proxies, while text and appearance references specify their intended final appearance.

\paragraph{Proxy-level construction.}
For Blender-authored scenes, the agent receives explicit instructions for each available fidelity level (\cref{fig:agent_construction}, step 3a). At \(Pr_0\), characters use coarse body blocks and root motion, with minimal arms or contact markers when essential to communicate an interaction. At \(Pr_1\), articulated heads, torsos, and limbs provide clearer poses and object interactions. At \(Pr_2\), additional geometry improves hands, feet, prop silhouettes, and local scene readability, together with fuller motion. Materials remain simple and untextured across these levels.

The execution harness runs the agent's generated \texttt{bpy} code in Blender and returns execution output, errors, scene information, and requested preview renders. The agent uses this feedback to construct and repair the scene incrementally before the final animation render. Where geometry-swap mode is used, a new tier starts from an existing animated proxy: replacement geometry inherits the original object's animation through a shared parent, while instructions preserve cameras, frame range, frame rate, timeline structure, role colours, and existing trajectories. These constraints guide the intended differences between levels; rendered inspection remains necessary to detect unintended changes. Native game and studio assets retain their available source representations and do not necessarily follow this Blender-authoring procedure.

\paragraph{Text-level construction and reference binding.}
The agent generates three levels of English prompts from the shared storyboard (\cref{fig:agent_construction}, step 3b). Each proxy-conditioned prompt begins with instructions on how to use the reference video, followed by a description of the intended final video. \(T_0\) provides a broad semantic and visual goal, relying mainly on the proxy for spatial and temporal structure. \(T_1\) adds appearance details and coarse descriptions of actions and shots. \(T_2\) provides detailed requirements for appearance, interactions, camera work, timing, and sound when applicable. We review the prompt content to maintain these distinctions and shorten overly detailed \(T_0\) prompts when necessary.

The proxy guides scene layout, camera motion, timing, and object movement, while text and appearance references specify characters, materials, environments, and visual style. Image and audio references are numbered according to their order in the input arrays. Input ablations retain the same target scene and remaining references, with prompts specifying the conditions available in each case. Appearance and voice requirements are expressed in text where needed in the absence of the corresponding reference media.

We use increasing character budgets as authoring guidelines: approximately 1,000--2,000 for \(T_0\), 2,500--3,500 for \(T_1\), and 4,000--5,000 for \(T_2\), including proxy and reference instructions. Prompt lengths reflect scene complexity and the information required at each level. The same finalized case prompt is used across supported models.

\paragraph{Optional voice-reference conversion.}
For some existing R2V cases with Chinese voice references, we automatically transcribe and translate the recordings into English, then synthesize English speech using voices cloned from the original recordings (\cref{fig:agent_construction}, step 3c). These translated recordings serve as voice references, not as dialogue that the generated video must reproduce.

The conversion targets speaker identity, intended meaning, and approximate speaking duration rather than exact word-level timing. Processed recordings are assigned to the corresponding numbered audio conditions and serve as references for generation and evaluation. We retain original recordings and conversion metadata where available and leave mixed-speaker or speech-with-music recordings unchanged.

\paragraph{Timeline verification with Blender sources.}
For audited scenes with available Blender files, we check explicit prompt timelines against both the source scene and the rendered proxy (\cref{fig:agent_construction}, step 4). Source inspection includes frame range and frame rate, camera assignments, timeline markers, relevant animation keyframes, and, where needed, sampled camera transforms and focal lengths. These provide supporting evidence for intended event timing, shot boundaries, and camera behavior. Frame indices are converted to seconds using the scene's starting frame and frame rate.

We cross-check source evidence with sampled video frames, including adjacent frames around proposed cuts and frames surrounding important actions or state changes. A timeline marker may describe an action rather than a cut, and a camera assignment need not produce a visible cut when successive views coincide. We therefore distinguish source annotations from events visible in the rendered video. When the evidence supports a continuous shot, the prompt describes successive action beats without introducing additional cuts.

These checks establish structure only to the extent supported by the proxy. Coarse geometry may show a trajectory or an object's broad location without establishing finger contact, subtle facial acting, or physical plausibility. Silent proxies cannot verify speech, knocks, or breathing. Such details remain target-generation instructions rather than claims about evidence present in the proxy.

\paragraph{Targeted correction and packaging.}
Corrections address specific observed defects and are followed by re-inspection (\cref{fig:agent_construction}, step 5). For example, in \emph{The key under the cup}, a hand in the intermediate proxy appeared too close to the cup during the reaction shot. A targeted repair instruction moved both hands away while preserving camera cuts, framing, role colours, and the cup and key trajectories. Later source and rendered-frame checks supported the cuts at 4 and 8 seconds. When the proxy agrees with the intended scene but an explicit prompt description conflicts with it, we revise the prompt to match the proxy.

We organize the corrected inputs into benchmark cases (\cref{fig:agent_construction}, step 6). For scenes with all three proxy levels, the five core configurations are \(Pr_0T_0\), \(Pr_0T_1\), \(Pr_0T_2\), \(Pr_1T_2\), and \(Pr_2T_2\), with additional variants where applicable. We also retain available Blender scenes, code, transcripts, metadata, previews, and earlier renderings to document the construction and revision process. The illustrated example has no image or audio conditions; the optional reference branches apply to other scenes.

\begin{table*}[t]
\centering
\caption{The 32 criteria and representative evidence.}
\label{tab:supp_criteria}
\footnotesize
\setlength{\tabcolsep}{3.5pt}
\renewcommand{\arraystretch}{1.12}
\begin{tabularx}{\textwidth}{@{}l>{\raggedright\arraybackslash}p{.245\textwidth}>{\raggedright\arraybackslash}p{.185\textwidth}@{\hspace{9pt}}l>{\raggedright\arraybackslash}p{.235\textwidth}>{\raggedright\arraybackslash}X@{}}
\toprule
\textbf{ID} & \textbf{Criterion} & \textbf{Evidence} & \textbf{ID} & \textbf{Criterion} & \textbf{Evidence} \\
\midrule
\multicolumn{3}{@{}l}{\textbf{P1 --- Spatial Structure}} & \multicolumn{3}{l@{}}{\textbf{C2 --- Motion \& Physical Plausibility}} \\
P1.1 & Entity presence and count & Counts and tracks & C2.1 & Kinematic and dynamic plausibility & Articulation; inertia \\
P1.2 & Projected position and scale & Boxes; screen positions & C2.2 & Temporal appearance coherence & Flicker; drift \\
P1.3 & Relative spatial relations & Timed relation checks & C2.3 & Contact and support plausibility & Contact intervals \\
P1.4 & Occlusion and depth order & Occlusion; depth cues & C2.4 & Dynamic geometry stability & Shape consistency \\
\addlinespace[4pt]
\multicolumn{3}{@{}l}{\textbf{P2 --- Motion \& Camera Control}} & \multicolumn{3}{l@{}}{\textbf{C3 --- Proxy Abstraction Handling}} \\
P2.1 & Coarse entity trajectory & Proxy-aligned tracks & C3.1 & Pseudo-action correction & Sliding; clipping; jumps \\
P2.2 & Direction and interaction phase & Action phases & C3.2 & Shape and structure completion & Target-object realization \\
P2.3 & Camera motion and shot plan & Viewpoints; shot cuts & C3.3 & Fine-grained motion completion & Missing articulation \\
P2.4 & Framing and composition & Subject scale; framing & C3.4 & Semantic detail completion & Specified semantic details \\
\addlinespace[4pt]
\multicolumn{3}{@{}l}{\textbf{P3 --- Temporal \& Shot Continuity}} & \multicolumn{3}{l@{}}{\textbf{B1 --- Entity \& Appearance Grounding}} \\
P3.1 & Event presence and order & Temporal event checks & B1.1 & Text-to-entity attributes & Entity-level attributes \\
P3.2 & Event timing and duration & Onsets; temporal windows & B1.2 & Image-to-entity appearance & Reference matching \\
P3.3 & Cross-shot action continuity & Action states across cuts & B1.3 & Cross-shot or cross-view identity & Identity across views \\
P3.4 & Persistent object state & Persistent state checks & B1.4 & Negative binding and leakage & Target / non-target checks \\
\addlinespace[4pt]
\multicolumn{3}{@{}l}{\textbf{C1 --- Visual Realization Quality}} & \multicolumn{3}{l@{}}{\textbf{B2 --- Action \& Audio-event Grounding}} \\
C1.1 & Realization completeness & Appearance of targets & B2.1 & Actor--action--target binding & Actor and target roles \\
C1.2 & Material, texture, and lighting & Material / light consistency & B2.2 & Sound-event correctness & Required sound / source \\
C1.3 & Perceptual and imaging quality & Image-quality cues & B2.3 & Voice and source assignment & Speaker / timbre matching \\
C1.4 & Local visual artifacts & Local distortions & B2.4 & Audio--visual synchronization & Event onset alignment \\
\bottomrule
\end{tabularx}
\end{table*}

\begin{table*}[t]
\centering
\caption{Checklist construction and verification.}
\label{tab:supp_prompt_pipeline}
\footnotesize
\setlength{\tabcolsep}{5pt}
\renewcommand{\arraystretch}{1.17}
\begin{tabularx}{\textwidth}{@{}p{.15\textwidth}p{.18\textwidth}p{.405\textwidth}>{\raggedright\arraybackslash}X@{}}
\toprule
\textbf{Stage} & \textbf{Input evidence} & \textbf{Procedure} & \textbf{Output} \\
\midrule
\textbf{Inventory} & Text and optional references & Identify named entities, attributes, actions, and sound events; retain concrete referents. & Target requirements for checklist questions. \\
\addlinespace[2pt]
\textbf{Proxy cues} & Proxy frames and inventory & Distinguish \emph{visible}, \emph{abstract stand-in}, and \emph{not observable}; unseen details are not automatically optional. & Structural evidence and completion targets. \\
\addlinespace[2pt]
\textbf{Questions} & Proxy, text, images/audio & Name the target entity or event; use text for intended content and the proxy for spatial and temporal structure. & Case-specific questions for P/C/B. \\
\addlinespace[2pt]
\textbf{Screening} & Proposed questions & Check answer form, polarity, grounding in available inputs, and unintended proxy-detail leakage. & Verified questions shared across models. \\
\addlinespace[2pt]
\textbf{Video judgment} & Questions and output video & Score with visual/audio evidence and the criterion rubric; subquestions do not receive separate weights. & Criterion scores for aggregation. \\
\bottomrule
\end{tabularx}
\end{table*}

\subsection{Video Generation Settings}
\label{sec:supp_generation_settings}
We evaluate Seedance-2.5, Kling-4.0, Wan-3.0, Seedance-2.0, and MiniMax-H3 using a common 720p resolution setting. Each scene specifies its target duration and framing, with 15-second videos and 16:9 or 21:9 aspect ratios as the standard formats. For each case, all supported models receive the same text prompt and reference conditions. The proxy is supplied directly as a reference video, with image, proxy, and audio inputs ordered consistently.

We retain one output per successfully generated model--case pair. Results are grouped by independent audio-reference availability, with Kling-4.0 excluded from the audio-reference group. Missing outputs receive zero, while unsupported configurations and cases that cannot be evaluated are excluded from the corresponding evaluation set.

\section{Evaluation Protocol}
\label{app:proxy_guidance}

\subsection{Checklist Construction}
\label{sec:supp_checklist}
For each case, Gemini-3.1-Pro instantiates the applicable criteria in Table~\ref{tab:supp_criteria} into questions grounded in the proxy, text, and available reference media. Qwen3.5-397B-A17B checks whether each question is clear, supported by visible inputs, and aligned with the intended criterion. Invalid questions are revised before video scoring, with deterministic checks used to normalize question scope and answer format. The resulting checklist is fixed and shared across models for that case. Multiple evidence subquestions contribute to a single criterion-level assessment without receiving separate weights.

\paragraph{Question formulation.}
Checklist construction starts from an inventory of the named entities, attributes,
actions, and events in the supplied conditions. The proxy contributes spatial and
temporal observations, while text and optional reference media specify the
intended appearance and assignments. Questions are written about concrete
entities and localized events rather than generic references to an unspecified
``action'' or ``object.'' Table~\ref{tab:supp_prompt_pipeline} summarizes the
information flow and verification requirements.

\begin{table*}[t]
\centering
\caption{Illustrative case-specific checklist covering all 32 criteria for a bakery dialogue scene. Questions are shortened and paraphrased for presentation rather than reproduced verbatim. Graded checks use the $0$--$4$ scale; trajectory matching uses a categorical answer. The remaining checks use binary verdicts.}
\label{tab:supp_full_checklist}
\footnotesize
\setlength{\tabcolsep}{4pt}
\renewcommand{\arraystretch}{1.12}
\begin{tabularx}{\textwidth}{@{}l>{\raggedright\arraybackslash}X@{\hspace{10pt}}l>{\raggedright\arraybackslash}X@{}}
\toprule
\textbf{ID} & \textbf{Case-specific question / requirement} & \textbf{ID} & \textbf{Case-specific question / requirement} \\
\midrule
\multicolumn{2}{@{}l}{\textbf{P1 --- Spatial Structure}} & \multicolumn{2}{l@{}}{\textbf{C2 --- Motion \& Physical Plausibility}} \\
P1.1 & Are one violinist, one baker, one parcel, one violin case, one table, two chairs, and the bakery fixtures present? & C2.1 & Rate the naturalness of the reach and the characters' head and torso motions. \\
P1.2 & Is the parcel in the lower-middle table area, occupying approximately 10--20\% of the frame width as in the proxy? & C2.2 & Rate temporal coherence of clothing and lighting across cuts. \\
P1.3 & Are the violinist on the left, the baker on the right, and the parcel between them, with stable scene anchors? & C2.3 & Does the baker's hand approach the parcel without clipping or implausible contact? \\
P1.4 & Are tabletop occlusion and foreground shoulder depth order consistent with the proxy? & C2.4 & Rate the stability of body proportions and hand geometry during the reach. \\
\addlinespace[3pt]
\multicolumn{2}{@{}l}{\textbf{P2 --- Motion \& Camera Control}} & \multicolumn{2}{l@{}}{\textbf{C3 --- Proxy Abstraction Handling}} \\
P2.1 & Is the seated violinist's coarse trajectory the same, mostly the same, or different from the proxy? & C3.1 & Rate the replacement of rigid proxy pivots and pseudo-actions with natural human motion. \\
P2.2 & Does the baker reach toward the parcel and then retract, following the proxy's interaction phases? & C3.2 & Rate completion of the characters, furniture, parcel, and other abstract proxy shapes. \\
P2.3 & Are the static two-shot and two reverse over-the-shoulder views preserved in order? & C3.3 & Rate fine articulation of faces, hands, and other motions omitted by the coarse proxy. \\
P2.4 & Are the intended shot sizes and overall composition of the table, characters, and background retained? & C3.4 & Rate the realization of clothing details, parcel wrapping, the violin case, and bakery fixtures. \\
\addlinespace[3pt]
\multicolumn{2}{@{}l}{\textbf{P3 --- Temporal \& Shot Continuity}} & \multicolumn{2}{l@{}}{\textbf{B1 --- Entity \& Appearance Grounding}} \\
P3.1 & Does the establishing conversation precede the parcel gesture and final reaction? & B1.1 & Are the green and cream outfits assigned to the violinist and baker, respectively? \\
P3.2 & Do the conversation (0--5 s), parcel gesture (5--10 s), and reaction (10--15 s) occur in their specified phases? & B1.2 & Does each character match the facial appearance and hair of their assigned image reference? \\
P3.3 & Does the reach/retract action remain continuous across corresponding cuts? & B1.3 & Does the violinist retain the assigned identity and appearance across shots? \\
P3.4 & Does the parcel remain intact, unopened, and stationary across all shots? & B1.4 & Are the specified clothing attributes kept from leaking to the wrong character? \\
\addlinespace[3pt]
\multicolumn{2}{@{}l}{\textbf{C1 --- Visual Realization Quality}} & \multicolumn{2}{l@{}}{\textbf{B2 --- Action \& Audio-event Grounding}} \\
C1.1 & Rate the completeness of the characters, props, and bakery setting, including natural faces and hands. & B2.1 & Does the baker, rather than the violinist, gesture toward the wrapped parcel? \\
C1.2 & Rate the plausibility of fabric, wrapping paper, wood materials, and soft window lighting. & B2.2 & Are the specified spoken dialogue phrases audible? \\
C1.3 & Rate focus, exposure, and freedom from imaging artifacts. & B2.3 & Are the feminine and masculine voice references assigned to the violinist and baker, respectively? \\
C1.4 & Rate rendering cleanliness, including the absence of clipping, melting, and corrupted textures. & B2.4 & Are the characters' lip movements synchronized with their respective dialogue lines? \\
\bottomrule
\end{tabularx}
\end{table*}

\paragraph{Proxy evidence.}
We construct a semantic inventory from the supplied text,
then assign each requirement one of three proxy-evidence states:
\emph{visible} when the relevant target is directly represented,
\emph{abstract stand-in} when a placeholder carries useful structure but not
final appearance or articulation, and \emph{not observable} when the proxy
cannot establish that detail. These states prevent missing proxy evidence
from being mistaken for permission to omit a text-specified entity or event.
They also clarify why following a placeholder trajectory belongs to Preserve,
whereas replacing unintended rigid articulation belongs to Complete.

\paragraph{Evaluator settings.}
Gemini-3.1-Pro serves as the primary visual and audio judge, using temperature $0$.
By default, the proxy and candidate timelines are presented side by side, together with uniformly sampled candidate frames (eight or twelve, depending on the capability), applicable reference images, and detail crops when needed. Requests use up to 64 visual inputs; audio-related judgments receive the candidate audio and relevant reference audio where required. The judge returns a structured answer, confidence, observed evidence, supporting timestamps, and criterion-specific subchecks. In the P/C/B evaluation pipeline, Qwen3.5-397B-A17B screens the checklist questions before video scoring.

\paragraph{Criterion boundaries.}
The categorization is determined by the \emph{requirement being tested}, not solely by the visual phenomenon. Preserve measures proxy-specified structure; Complete measures realization of unspecified content and correction of unintended proxy abstraction; Bind measures assignment of text, image, and audio conditions to the designated entities and events. For example, P1.1 checks the number of entities, whereas C1.1 checks their realized appearance. Similarly, P2 checks whether a coarse proxy trajectory is respected, C2 judges the naturalness of the motion, and B2 checks whether an action is assigned to the specified actor. This limits duplicate scoring of one defect under multiple labels.

\paragraph{Checklist validity.}
We audit input support at the level of instantiated assessments across evaluated outputs, rather than counting unique criterion definitions. Among 41,067 check instances, 50 were flagged for applicability review, and none remained unsupported after input-dependent filtering. This assessment addresses question grounding, complementing the human validation of video-level judgments in Sec.~\ref{sec:supp_human}.

\paragraph{Full checklist example.}
Table~\ref{tab:supp_full_checklist} illustrates how the 32 criteria are instantiated for a bakery dialogue scene with two seated characters, a wrapped parcel, a three-shot sequence, and image and voice references. Each question specifies the entities, events, or realization requirements to be assessed. Only applicable criteria enter the case-level evaluation.

\subsection{Criterion Scoring and Applicability}
\label{sec:supp_criterion_scoring}
Binary checks receive 1 if the expected answer is established and 0 otherwise. For P2.1, both \emph{same} and \emph{mostly same} count as a pass. Graded Complete checks use criterion-specific integer rubrics from 0 (not realized or severe defects) to 4 (fully realized with no relevant defects), normalized by dividing the level by 4. Compound binary requirements remain one score; a failed required subcheck overrides an affirmative overall verdict. Preserve, Bind, and the binary contact check C2.3 use a pass threshold of 0.50; the other Complete criteria use 0.60. Inapplicable criteria do not contribute to capability or dimension scores.

\paragraph{Cross-configuration evaluation.}
Matched configurations share a common criterion taxonomy, with questions instantiated from the available inputs. Preserve and Complete assess structural adherence and content realization, while additional image and audio references introduce grounding requirements under B1 and B2, respectively. Reference-specific binding is assessed on the corresponding applicable criteria.

\subsection{Score Aggregation}
\label{sec:supp_scores}
Let $s_{i,k}\in[0,1]$ be the score for criterion $k$ in case $i$, and let $\mathcal K_{i,g}$ be the applicable criteria for capability $g$. With the fixed criterion weights $a_k=1$, the capability score is
\begin{equation}
 f_{i,g}=\frac{\sum_{k\in\mathcal K_{i,g}}a_k s_{i,k}}{\sum_{k\in\mathcal K_{i,g}}a_k}.
\end{equation}
For dimension $d$, let $\mathcal G_{i,d}$ denote its applicable capabilities. The dimension score averages these capability scores with equal weights:
\begin{equation}
 D_{i,d}=\frac{1}{|\mathcal G_{i,d}|}\sum_{g\in\mathcal G_{i,d}} f_{i,g}.
\end{equation}
The case-level mean and structure-sensitive scores are
\begin{align}
 Q_i &= \frac{1}{|\mathcal D_i|}\sum_{d\in\mathcal D_i}D_{i,d},\\
 \rho_i &= \frac{1}{12}\sum_{k\in\mathcal P}\mathbf 1[\mathrm{pass}(i,k)],\\
 S_i &= Q_i\rho_i^2.
\end{align}
Here, $\mathcal D_i$ contains the applicable dimensions and $\mathcal P=\{\mathrm{P1.1},\ldots,\mathrm{P3.4}\}$ is the fixed set of 12 critical Preserve criteria. The plans for all 294 proxy-conditioned cases support these 12 criteria, giving an effective denominator of 12 for scored proxy-conditioned outputs. The 18 no-proxy counterparts are evaluated separately using the control metrics below. \emph{Joint Structural Success} is $\mathbf1[\rho_i=1]$; it is not a joint pass requirement over P/C/B.

\subsection{Control Metrics}
\label{sec:supp_proxy_control}
To measure the benefits of structural guidance, we evaluate 50 matched proxy-conditioned and no-proxy video pairs from 18 scenes, with a shared target scene description, other references, and generation settings. We assess action transitions, cross-shot spatial consistency, and multi-view geometric consistency using complementary evaluation protocols.

\paragraph{Action Transition.}
We adopt the temporal state-transition and ordering checks of TC-Bench and the MLLM-based action-binding and object-interaction evaluation of T2V-CompBench, adapting them to our dynamic-action scenes. Each scene contains 2--4 predefined binary assertions covering action occurrence, state-transition completion, and temporal order. Qwen3.5-397B-A17B evaluates each generated video against the same assertions, returning a pass/fail judgment for each requirement.

\paragraph{Cross-shot Spatial Consistency.}
We adopt the relational evaluation of T2V-CompBench and the cross-shot continuity checks of MSVBench, adapting them to multi-shot scenes. For each scene, we define fixed assertions concerning persistent entities, landmarks, object states, and spatial relationships across shots. Using predefined shot boundaries, we sample one timestamped frame per shot. Qwen3.5-397B-A17B evaluates these frames in temporal order, allowing viewpoint changes while checking the spatial topology. 

\begin{table}[t]
\centering
\caption{Fidelity effects ($Pr_2-Pr_0$) on 31 matched scenes.}
\label{tab:app_fidelity_audio}
\footnotesize
\setlength{\tabcolsep}{4pt}
\renewcommand{\arraystretch}{1.17}
\begin{tabular*}{\columnwidth}{@{\extracolsep{\fill}}lrrr@{}}
\toprule
\textbf{Model} & \textbf{$\Delta P$} & \textbf{$\Delta C$} & \textbf{$\Delta B$} \\
\midrule
Seedance-2.5 & +0.022 & -0.060 & +0.040 \\
Kling-4.0 & +0.030 & -0.072 & -0.020 \\
Wan-3.0 & +0.011 & -0.056 & -0.036 \\
Seedance-2.0 & +0.016 & -0.048 & -0.089 \\
MiniMax-H3 & +0.067 & -0.079 & -0.182 \\
\midrule
\textbf{Mean} & \textbf{+0.029} & \textbf{-0.063} & \textbf{-0.057} \\
\bottomrule
\end{tabular*}
\end{table}

\paragraph{Assertion-based Scoring.}
For either metric $m$, let $\mathcal{A}_s^m$ denote the predefined assertions for scene $s$, and $J(v,a)\in\{0,1\}$ indicate whether video $v$ satisfies assertion $a$. The video-level score and paired improvement are
\begin{equation}
\begin{aligned}
q_m(v,s) &= \frac{1}{|\mathcal{A}_s^m|}
\sum_{a\in\mathcal{A}_s^m} J(v,a),\\
\Delta_m &= \frac{1}{N_m}\sum_{i=1}^{N_m}
\left(q_m(v_i^+,s_i)-q_m(v_i^-,s_i)\right),
\end{aligned}
\label{eq:supp_control_score}
\end{equation}
where $v_i^+$ and $v_i^-$ denote proxy-conditioned and no-proxy outputs, respectively, and $N_m$ is the number of valid matched pairs. Thus, assertions are averaged within each video before computing the mean paired difference.

\paragraph{Multi-view Consistency.}
We adopt the official MEt3R metric to evaluate geometric consistency across overlapping views. MEt3R compares corresponding scene content using estimated geometry and visual features without requiring ground-truth camera poses. We apply it to valid overlapping view pairs, retaining dynamic content, and aggregate the resulting errors. Unlike the assertion-based metrics, lower MEt3R scores indicate better consistency.

\paragraph{Pairwise VLM Preference.}
We additionally conduct holistic VLM comparisons on all 50 matched pairs. Each pair is evaluated in both presentation orders to mitigate positional bias. The primary order yields 40 proxy wins, two ties, and eight losses. Combining both orders produces 35 consistent proxy wins and 15 ties or order-dependent outcomes, with no consistent proxy losses.

The action-transition, cross-shot spatial, and MEt3R evaluations include 22 pairs from eight scenes, 14 pairs from five scenes, and 17 pairs from ten scenes, respectively. These metric-valid subsets may overlap, while VLM preference covers all 50 matched pairs. Together, the measures quantify complementary aspects of the structural benefits reported in main Table~2.

\section{Proxy Fidelity}
\label{sec:app_fidelity}
\subsection{Matched Fidelity Trends}
We analyze the effect of proxy fidelity on 31 matched scenes, with all three proxy levels ($Pr_0$, $Pr_1$, and $Pr_2$), fixed $T_2$ text, and no independent audio references. Table~\ref{tab:app_fidelity_audio} reports the model-level changes from $Pr_0$ to $Pr_2$.
All five models exhibit higher mean Preserve and lower mean Complete at $Pr_2$ than at $Pr_0$, while Bind responds differently across models. Notably, Seedance-2.5 gains 0.040 in Bind, whereas MiniMax-H3 loses 0.182, indicating that stronger proxy guidance does not necessarily facilitate reference grounding.

The intermediate fidelity level reveals a further distinction. Averaged equally across models, Preserve increases from 0.738 at $Pr_0$ to 0.776 at $Pr_1$, but slightly decreases to 0.767 at $Pr_2$. In contrast, Complete declines monotonically from 0.601 to 0.554 and 0.538. Most of the completion loss therefore occurs between $Pr_0$ and $Pr_1$ (-0.047), with a smaller additional decline from $Pr_1$ to $Pr_2$ (-0.016). The corresponding average scores $Q=(0.696, 0.668, 0.665)$ show that additional proxy detail does not translate into higher overall performance. These results suggest diminishing structural benefits alongside persistent completion costs, rather than a uniformly beneficial increase in proxy fidelity.

\begin{figure*}[t]
    \centering
    \includegraphics[width=.88\textwidth]{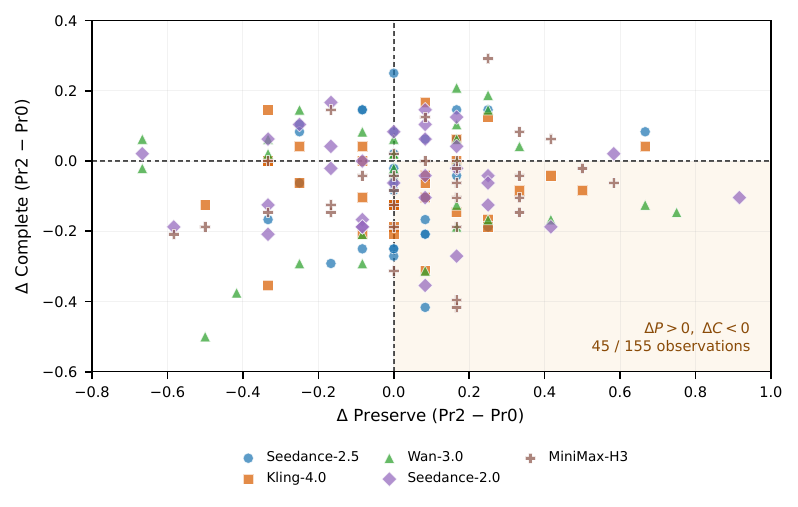}
    \caption{Scene-level Preserve and Complete changes from $Pr_0$ to $Pr_2$ (155 matched observations).}
    \label{fig:supp_pc_tradeoff}
\end{figure*}

\subsection{Scene-Level Variation}
\label{sec:supp_fidelity_hetero}
To examine whether the fidelity trade-off holds across individual scenes, we analyze 155 matched model--scene pairs from the same 31 scenes. Table~\ref{tab:supp_fidelity_hetero} and Fig.~\ref{fig:supp_pc_tradeoff} summarize the joint changes in Preserve and Complete from $Pr_0$ to $Pr_2$.
Notably, 49 of 155 pairs (31.6\%) exhibit reduced Complete without improved Preserve, indicating that additional proxy detail can incur completion costs without corresponding structural benefits. Conversely, 29 pairs (18.7\%) improve Preserve without sacrificing Complete. Scene-cluster bootstrap estimates further show that the Complete decline ($\Delta C=-0.063$, 95\% CI $[-0.082,-0.043]$) is more consistent than the Preserve improvement ($\Delta P=+0.029$, 95\% CI $[-0.030,+0.095]$). These results highlight the importance of selectively exploiting proxy guidance rather than uniformly following additional proxy detail.

\begin{table}[t]
\centering
\caption{Joint Preserve and Complete changes across 155 matched model--scene pairs.}
\label{tab:supp_fidelity_hetero}
\footnotesize
\setlength{\tabcolsep}{5pt}
\renewcommand{\arraystretch}{1.18}
\begin{tabular*}{\columnwidth}{@{\extracolsep{\fill}}lrrr@{}}
\toprule
& $\boldsymbol{\Delta C<0}$ & $\boldsymbol{\Delta C\geq 0}$ & \textbf{Total} \\
\midrule
$\Delta P>0$ & 45 & 29 & 74 \\
$\Delta P\leq 0$ & 49 & 32 & 81 \\
\midrule
\textbf{Total} & \textbf{94} & \textbf{61} & \textbf{155} \\
\bottomrule
\end{tabular*}
\end{table}

\subsection{Completion Bottleneck}
\label{sec:supp_capability_fidelity}

To identify which capabilities drive the preservation--completion tension, we further decompose the fidelity results into eight capabilities. Using the same 31 matched $T_2$ scenes and five models, Table~\ref{tab:supp_capability_fidelity} reports the model-averaged changes from $Pr_0$ to $Pr_2$, together with scene-cluster bootstrap confidence intervals and the direction of change across scenes.

\begin{table*}[t]
\centering
\caption{Capability-level fidelity changes ($Pr_2-Pr_0$). All capabilities are evaluated on 31 matched scenes except B2 ($n=21$). Signs count scene-level model-averaged changes; confidence intervals are obtained by scene-cluster bootstrap.}
\label{tab:supp_capability_fidelity}
\footnotesize
\setlength{\tabcolsep}{5pt}
\renewcommand{\arraystretch}{1.15}
\begin{tabularx}{\textwidth}{@{}p{.235\textwidth}r r c >{\raggedright\arraybackslash}X@{}}
\toprule
\textbf{Capability} & \textbf{Mean $\Delta$} & \textbf{95\% CI} & \textbf{Scenes $+/-/0$} & \textbf{Evaluation focus} \\
\midrule
P1 Spatial Structure & +0.034 & [-0.048, +0.113] & 15/12/4 & Entity placement and spatial relationships \\
P2 Motion \& Camera & +0.024 & [-0.050, +0.108] & 15/10/6 & Coarse trajectories and camera control \\
P3 Temporal Continuity & +0.029 & [-0.044, +0.110] & 16/13/2 & Event order and cross-shot continuity \\
\addlinespace[3pt]
\midrule
C1 Visual Realization & -0.036 & [-0.073, -0.002] & 12/17/2 & Appearance, texture, and visual coherence \\
C2 Motion \& Physics & -0.048 & [-0.102, +0.012] & 8/21/2 & Fine-grained motion and physical interactions \\
\textbf{C3 Proxy Abstraction} & \textbf{-0.104} & \textbf{[-0.151, -0.057]} & \textbf{7/24/0} & \textbf{Artifact correction and detail completion} \\
\addlinespace[3pt]
\midrule
B1 Entity \& Appearance & -0.060 & [-0.134, +0.005] & 7/10/14 & Entity identity and appearance grounding \\
B2 Action \& Audio-event & +0.010 & [-0.171, +0.200] & 8/9/4 & Action, voice, and event grounding \\
\bottomrule
\end{tabularx}
\end{table*}

The completion decline is dominated by C3 (Proxy Abstraction Handling), which decreases by 0.104 (95\% CI: [-0.151, -0.057]) and shows negative changes in 24 of 31 scenes. Its decline is more than twice that of C2 (-0.048) and accounts for approximately 55\% of the summed decreases across the three Complete capabilities. C1 also exhibits a smaller but consistent decline (-0.036), whereas the uncertainty interval for C2 overlaps zero.

In contrast, all three Preserve capabilities show positive mean changes but substantial scene-level variability, with confidence intervals spanning zero. This reveals an asymmetry in how models respond to increased proxy detail: structural adherence may improve, but the ability to replace proxy abstractions with completed appearance and motion becomes more challenging. The particularly large decline in C3 suggests that the main difficulty lies in distinguishing useful structural guidance from proxy-specific details that should be refined, rather than a uniform deterioration in visual or motion quality.

\section{Condition and Binding Analysis}
\subsection{Input Richness and Binding}
\label{sec:supp_cross_conditions}
Structural proxies provide spatial and temporal guidance, while text and additional references specify the desired realization and the assignment of attributes and events. Increasing input richness therefore provides more precise guidance while introducing additional grounding requirements. Table~\ref{tab:input_condition_details} summarizes the configuration-level Strict differences, complemented by fine-grained binding diagnostics.

\begin{table*}[t]
\centering
\caption{Input richness and reference grounding. $\Delta$Strict denotes model-averaged configuration-level differences in percentage points. Fine-grained binding scores are evaluated on their applicable subsets.}
\label{tab:input_condition_details}
\footnotesize
\setlength{\tabcolsep}{5pt}
\renewcommand{\arraystretch}{1.18}
\begin{tabularx}{\textwidth}{@{}p{.20\textwidth}c r p{.16\textwidth}>{\raggedright\arraybackslash}X@{}}
\toprule
\textbf{Contrast} & \textbf{Models} & \textbf{$\Delta$Strict (pp)} & \textbf{Range (pp)} & \textbf{Fine-grained diagnostic} \\
\midrule
$Pr_0$: $T_0\rightarrow T_2$ & 5 & +5.2 & +2.4 to +7.7 & More detailed appearance, interactions, and event timing. \\
\addlinespace[3pt]
$Pr{+}T\rightarrow {+}I$ & 5 & -10.4 & -16.6 to -6.3 & Identity: 0.897; cross-shot identity: 0.909; action--object: 0.636. \\
$Pr{+}T\rightarrow {+}A$ & 4 & -13.8 & -23.2 to -8.9 & Voice/source: 0.768; audiovisual timing: 0.513. \\
$Pr{+}T\rightarrow {+}I{+}A$ & 4 & -15.2 & -24.0 to -4.4 & Joint identity, appearance, and audio-event grounding. \\
\bottomrule
\end{tabularx}
\end{table*}

\begin{table*}[t]
\centering
\caption{Paired targeted-case comparisons. Original denotes the matched same-scene $Pr_0$--$T_2$ configuration, except for voice swaps, where it denotes the original speaker assignment. Scores are averaged over models with both outputs. $\Delta$ is Variant minus Original, calculated before rounding; $N$ denotes the number of paired models (4 for reference-removal and voice cases, 5 otherwise).}
\label{tab:supp_targeted}
\footnotesize
\setlength{\tabcolsep}{5pt}
\renewcommand{\arraystretch}{1.14}
\begin{tabular*}{\textwidth}{@{\extracolsep{\fill}}lrrrrrrr@{}}
\toprule
& \multicolumn{3}{c}{\textbf{Preserve}} & \multicolumn{3}{c}{\textbf{Bind}} & \\
\cmidrule(lr){2-4}
\cmidrule(lr){5-7}
\textbf{Variation} & \textbf{Orig.} & \textbf{Var.} & $\boldsymbol{\Delta P}$ & \textbf{Orig.} & \textbf{Var.} & $\boldsymbol{\Delta B}$ & $\boldsymbol{N}$ \\
\midrule
Joint image/audio removal A & 0.625 & 0.688 & +0.062 & 0.844 & 0.750 & -0.094 & 4 \\
Joint image/audio removal B & 0.625 & 0.500 & -0.125 & 0.594 & 0.250 & -0.344 & 4 \\
Joint image/audio removal C & 0.583 & 0.771 & +0.188 & 0.583 & 1.000 & +0.417 & 4 \\
\addlinespace[4pt]
Voice order swap A & 0.625 & 0.667 & +0.042 & 0.812 & 0.656 & -0.156 & 4 \\
Voice order swap B & 0.854 & 0.833 & -0.021 & 0.844 & 0.844 & 0.000 & 4 \\
\addlinespace[4pt]
Identity swap & 0.833 & 0.800 & -0.033 & 0.850 & 0.500 & -0.350 & 5 \\
Image assignment swap & 0.950 & 0.817 & -0.133 & 0.925 & 0.900 & -0.025 & 5 \\
\addlinespace[4pt]
Delayed gate & 0.783 & 0.767 & -0.017 & 0.600 & 0.700 & +0.100 & 5 \\
Dolly only & 0.750 & 0.950 & +0.200 & 1.000 & 0.600 & -0.400 & 5 \\
Zoom only & 0.750 & 0.867 & +0.117 & 1.000 & 0.900 & -0.100 & 5 \\
Early focus & 0.800 & 0.683 & -0.117 & 1.000 & 0.900 & -0.100 & 5 \\
Centered camera & 0.833 & 0.933 & +0.100 & 0.800 & 1.000 & +0.200 & 5 \\
\bottomrule
\end{tabular*}
\end{table*}

\paragraph{Text as complementary guidance.}
At fixed $Pr_0$, moving from $T_0$ to $T_2$ yields higher observed Strict scores for all five models, with gains of 2.4--7.7 percentage points. Detailed text supplies appearance, interaction, and temporal requirements beyond the coarse proxy structure. However, Seedance-2.0 achieves its highest text-level Strict at $T_1$, indicating that the benefit depends on the model rather than increasing monotonically with prompt detail.

\paragraph{Reference grounding.}
Reference-conditioned generation reveals a distinction between recognizing reference content and grounding it in the intended scene. Identity matching (0.897) and cross-shot identity consistency (0.909) exceed action--object binding (0.636), while voice/source matching (0.768) exceeds audiovisual timing alignment (0.513). These correspond to gaps of 26.1 and 25.5 percentage points, respectively. Thus, reproducing reference appearance or sound does not necessarily ensure its correct assignment to the intended interaction or temporal event.

The lower configuration-level Strict scores under additional image or audio conditioning reflect the combined demands of structural preservation, content completion, and newly specified binding requirements. Together, these results highlight the challenge of coordinating multiple conditions within a fixed proxy-defined scene.

\subsection{Targeted Cases}
\label{sec:supp_targeted_cases}
We examine 14 targeted cases through 12 paired comparisons involving reference removal, identity and voice assignment, event timing, and camera control (Table~\ref{tab:supp_targeted}). The two voice-assignment comparisons each combine original and swapped configurations. These controlled variations reveal that structural preservation and condition binding can respond differently to changes in input requirements.

Identity swapping reduces Bind by 0.350 while Preserve decreases by only 0.033. Similarly, exchanging voice references in Dialogue A decreases Bind by 0.156 despite a 0.042 increase in Preserve, whereas Dialogue B maintains the same Bind score under both assignments. Camera-control variations also exhibit divergent responses: the dolly-only case improves Preserve by 0.200 while Bind decreases by 0.400. Reference removal produces heterogeneous configuration-level Bind changes, ranging from -0.344 to +0.417, as the available grounding requirements vary across scenes.

These paired comparisons demonstrate that preserving proxy-defined structure does not guarantee correct grounding of modified input conditions. The response to individual control changes also varies across scenes and requirement types, further motivating the separate evaluation of Preserve and Bind.

\begin{table*}[t]
\centering
\caption{Common-support performance with scene-cluster bootstrap 95\% confidence intervals. Models within each input group are evaluated on identical cases.}
\label{tab:supp_common_support}
\footnotesize
\setlength{\tabcolsep}{7pt}
\renewcommand{\arraystretch}{1.15}
\begin{tabular*}{\textwidth}{@{\extracolsep{\fill}}lrc rc@{}}
\toprule
\textbf{Model} & \textbf{$Q$} & \textbf{95\% CI ($Q$)} & \textbf{Strict} & \textbf{95\% CI (Strict)} \\
\midrule
\multicolumn{5}{@{}l}{\textit{Without independent audio (190 cases, 48 scenes)}} \\
\addlinespace[2pt]
Seedance-2.5 & 0.692 & [0.663, 0.720] & 0.464 & [0.414, 0.514] \\
Kling-4.0 & 0.681 & [0.650, 0.712] & 0.442 & [0.395, 0.487] \\
Wan-3.0 & 0.658 & [0.626, 0.688] & 0.415 & [0.370, 0.459] \\
Seedance-2.0 & 0.636 & [0.600, 0.671] & 0.395 & [0.339, 0.451] \\
MiniMax-H3 & 0.655 & [0.624, 0.686] & 0.370 & [0.321, 0.419] \\
\addlinespace[4pt]
\midrule
\multicolumn{5}{@{}l}{\textit{With independent audio (44 cases, 9 scenes)}} \\
\addlinespace[2pt]
Seedance-2.5 & 0.659 & [0.607, 0.707] & 0.303 & [0.232, 0.374] \\
Wan-3.0 & 0.651 & [0.618, 0.686] & 0.364 & [0.297, 0.424] \\
Seedance-2.0 & 0.626 & [0.576, 0.671] & 0.293 & [0.219, 0.370] \\
MiniMax-H3 & 0.601 & [0.555, 0.644] & 0.256 & [0.212, 0.294] \\
\bottomrule
\end{tabular*}
\end{table*}

\begin{table*}[t]
\centering
\caption{Selected paired Strict comparisons on common-support cases. Differences are computed before rounding. Confidence intervals are pointwise; $p$-values are Holm-adjusted across all model pairs in each group.}
\label{tab:app_total_models}
\footnotesize
\setlength{\tabcolsep}{5pt}
\renewcommand{\arraystretch}{1.16}
\begin{tabularx}{\textwidth}{@{}p{.30\textwidth}r r c>{\raggedright\arraybackslash}X@{}}
\toprule
\textbf{Model pair (first $-$ second)} & \textbf{$\Delta S$} & \textbf{95\% CI} & \textbf{Holm $p$} & \textbf{Finding} \\
\midrule
\multicolumn{5}{@{}l}{\textit{Without independent audio}} \\
\addlinespace[2pt]
Seedance-2.5 $-$ Kling-4.0 & +0.023 & [-0.033, 0.076] & 0.935 & Top-two separation is uncertain. \\
Seedance-2.5 $-$ Seedance-2.0 & +0.069 & [0.024, 0.113] & 0.056 & Positive difference, but not significant after correction. \\
Seedance-2.5 $-$ MiniMax-H3 & +0.095 & [0.057, 0.133] & $<0.001$ & Significant structural performance gap. \\
\addlinespace[3pt]
\midrule
\multicolumn{5}{@{}l}{\textit{With independent audio}} \\
\addlinespace[2pt]
Seedance-2.5 $-$ Wan-3.0 & -0.061 & [-0.146, 0.030] & 0.980 & No reliable separation. \\
Wan-3.0 $-$ MiniMax-H3 & +0.108 & [0.035, 0.172] & 0.199 & Positive estimate, but not significant after correction. \\
\bottomrule
\end{tabularx}
\end{table*}

\section{Statistical Validation}
\label{sec:app_stats}

\subsection{Statistical Protocol}
\label{sec:supp_stats_protocol}
We use scenes as the resampling unit to account for correlations among multiple configurations and model outputs derived from the same scene. Confidence intervals are estimated using 5,000 scene-cluster bootstrap replicates. Common-support comparisons retain the intersection of cases with scored outputs for every model within the corresponding input group, ensuring identical scene conditions and checklists.
The common-support sets contain 190 cases from 48 scenes without independent audio and 44 cases from nine scenes with independent audio. Table~\ref{tab:supp_common_support} reports the corresponding results. These shared-case estimates complement the broader model comparisons in the main manuscript.

The common-support results reveal differences between average performance and structural reliability. Without independent audio, MiniMax-H3 achieves a higher $Q$ than Seedance-2.0 (0.655 versus 0.636), but a lower Strict score (0.370 versus 0.395). With audio, Wan-3.0 achieves a higher Strict score than Seedance-2.5 (0.364 versus 0.303), despite a slightly lower $Q$. These differences illustrate how joint structural constraints can distinguish models beyond their average dimension scores.

\subsection{Paired Comparisons}
\label{sec:supp_pairwise}
We further assess model differences using paired scene-cluster sign-flip tests on the common-support sets. Holm correction is applied across all model pairs within each input group. Table~\ref{tab:app_total_models} reports representative comparisons with pointwise 95\% confidence intervals and adjusted $p$-values.

The results indicate that some structural-performance differences are more robust than others. Without independent audio, Seedance-2.5 significantly outperforms MiniMax-H3 on Strict after Holm correction, whereas the gap between Seedance-2.5 and Kling-4.0 is not statistically significant. None of the audio-group pairwise differences survives correction. Thus, the benchmark reveals meaningful model-level differences while also distinguishing stable separations from closely matched performance.

\subsection{Strict Sensitivity}
\label{sec:app_strict}
To examine the influence of structural weighting, we vary the exponent in the Strict score while keeping the underlying criterion judgments fixed:
\begin{equation}
S_i^{(\alpha)}=Q_i\rho_i^{\alpha},\qquad \alpha\in\{0,1,2,3\}.
\end{equation}
The primary definition uses $\alpha=2$, while $\alpha=0$ corresponds to the average dimension score $Q$. Table~\ref{tab:supp_strict_sensitivity} reports results on the same input-group populations as the main benchmark.

\begin{table}[t]
\centering
\caption{Sensitivity to structural weighting ($\alpha=2$ is the primary setting).}
\label{tab:supp_strict_sensitivity}
\footnotesize
\setlength{\tabcolsep}{4pt}
\renewcommand{\arraystretch}{1.15}
\begin{tabular*}{\columnwidth}{@{\extracolsep{\fill}}lrrrr@{}}
\toprule
\textbf{Model} & $\alpha=0$ & $1$ & \textbf{$2$} & $3$ \\
\midrule
\multicolumn{5}{@{}l}{\textit{Without independent audio}} \\
Seedance-2.5 & 0.678 & 0.538 & \textbf{0.450} & 0.387 \\
Kling-4.0 & 0.663 & 0.517 & \textbf{0.425} & 0.362 \\
Wan-3.0 & 0.644 & 0.488 & \textbf{0.399} & 0.340 \\
Seedance-2.0 & 0.615 & 0.459 & \textbf{0.375} & 0.321 \\
MiniMax-H3 & 0.615 & 0.439 & \textbf{0.344} & 0.283 \\
\addlinespace[3pt]
\midrule
\multicolumn{5}{@{}l}{\textit{With independent audio}} \\
Seedance-2.5 & 0.666 & 0.446 & \textbf{0.325} & 0.251 \\
Wan-3.0 & 0.584 & 0.416 & \textbf{0.324} & 0.265 \\
Seedance-2.0 & 0.634 & 0.421 & \textbf{0.314} & 0.252 \\
MiniMax-H3 & 0.581 & 0.350 & \textbf{0.241} & 0.179 \\
\bottomrule
\end{tabular*}
\end{table}

The no-audio ranking remains stable across the tested exponents, apart from a tie at reported precision for $\alpha=0$. In contrast, the audio-group ranking changes as structural penalties become stronger: Wan-3.0 overtakes Seedance-2.5 at $\alpha=3$. This indicates that Strict emphasizes a distinct aspect of model performance, while its effect on ranking depends on the distribution of joint structural successes. Reporting P/C/B alongside Strict therefore preserves both diagnostic detail and structure-sensitive comparison.

\subsection{Human Validation}
\label{sec:supp_human}
The main human evaluation covers 3,199 criterion-level judgments from 110 videos, with an overall agreement of 92.9\% (89.7\% for Preserve, 95.2\% for Complete, and 94.8\% for Bind). To further identify challenging evaluation capabilities, we examine 611 judge-labeled failures at the capability level (Table~\ref{tab:app_audit}).

\begin{table}[t]
\centering
\caption{Capability-level human agreement on judge-labeled failures, with Wilson 95\% confidence intervals.}
\label{tab:app_audit}
\footnotesize
\setlength{\tabcolsep}{3.5pt}
\renewcommand{\arraystretch}{1.16}
\begin{tabular*}{\columnwidth}{@{\extracolsep{\fill}}lrrc@{}}
\toprule
\textbf{Capability} & \textbf{Fail $N$} & \textbf{Agreement} & \textbf{95\% CI} \\
\midrule
P1 Spatial structure & 77 & 0.844 & [0.747, 0.909] \\
P2 Motion / camera & 46 & 0.935 & [0.825, 0.978] \\
P3 Temporal continuity & 47 & 0.830 & [0.699, 0.911] \\
\addlinespace[3pt]
\midrule
C1 Visual realization & 71 & 0.986 & [0.924, 0.998] \\
C2 Motion / physics & 120 & 0.858 & [0.785, 0.910] \\
C3 Proxy abstraction & 119 & 0.933 & [0.873, 0.966] \\
\addlinespace[3pt]
\midrule
B1 Entity / appearance & 96 & 0.885 & [0.806, 0.935] \\
\textbf{B2 Action / audio event} & \textbf{35} & \textbf{0.714} & \textbf{[0.549, 0.837]} \\
\bottomrule
\end{tabular*}
\end{table}

Human agreement on judge-labeled failures varies across capabilities. It is lowest for B2 (0.714), followed by P3 (0.830) and P1 (0.844), whereas B1 achieves 0.885. The results suggest that assessing event timing, sound-source assignments, and actor--action relationships is more challenging than assessing entity appearance, highlighting the importance of localized temporal and relational evidence for reliable evaluation.

\section{Limitations and Future Work}
\label{sec:supp_limitations_impacts}
\paragraph{Limitations.}
Our proxy fidelity levels jointly vary geometric and motion detail, making it difficult to isolate their individual effects. Moreover, evaluating subtle physical interactions and precise audiovisual synchronization remains challenging for current VLM-based evaluators.

\paragraph{Future Work.}
Our findings motivate more effective proxy-guided generation methods that selectively preserve intended structure while refining incomplete geometry and motion. Future research could explore adaptive proxy conditioning and proxy-aware training to better balance Preserve and Complete, alongside stronger entity- and event-level multimodal grounding to improve Bind.

\end{document}